\documentclass[10pt,twocolumn,letterpaper]{article}
\pdfoutput=1
\usepackage{cvpr}              

\usepackage{graphicx}
\usepackage{amsmath}
\usepackage{amssymb}
\usepackage{booktabs}%
\usepackage{cuted}
\usepackage{capt-of}
\usepackage{caption}

\usepackage{amsmath,amsfonts,bm}
\usepackage{multirow}
\usepackage{tabularx}
\usepackage{graphicx}
\usepackage{adjustbox}
\usepackage{amsthm}
\usepackage{xspace}

\renewcommand{\arraystretch}{1.3}
\newcommand{\mset}[1]{\left\{\kern-.5em\left\{ #1 \right\}\kern-.5em\right\}}
\newcommand{\mmset}[1]{\{\kern-.4em\{ #1 \}\kern-.4em\}}

\newcommand{\parr}[1]{\left (#1\right )}

\def\eqref#1{equation~\ref{#1}}

\def\1{\bm{1}}

\def\vc{{\bm{c}}}
\def\vd{{\bm{d}}}

\def\vf{{\bm{f}}}

\def\vl{{\bm{l}}}

\def\vo{{\bm{o}}}
\def\vp{{\bm{p}}}

\def\vr{{\bm{r}}}

\def\vw{{\bm{w}}}
\def\vx{{\bm{x}}}

\def\vz{{\bm{z}}}

\DeclareMathAlphabet{\mathsfit}{\encodingdefault}{\sfdefault}{m}{sl}
\SetMathAlphabet{\mathsfit}{bold}{\encodingdefault}{\sfdefault}{bx}{n}

\usepackage{diagbox} 

\usepackage{multirow}
\usepackage{booktabs} 

\usepackage{threeparttable}
\usepackage{pifont}
\usepackage{enumitem}
\usepackage{color}
\usepackage{xcolor}  
\newcommand{\minisection}[1]{\vspace{0.04in} \noindent {\bf #1}\ \ }

\usepackage[pagebackref,breaklinks,colorlinks]{hyperref}

\usepackage{caption}

\usepackage[capitalize]{cleveref}
\crefname{section}{Sec.}{Secs.}
\Crefname{section}{Section}{Sections}
\Crefname{table}{Table}{Tables}
\crefname{table}{Tab.}{Tabs.}

\def\cvprPaperID{81} 
\def\confName{CVPR}
\def\confYear{2023}

\begin{document}

\title{3D-Aware Multi-Class Image-to-Image Translation with NeRFs}

\author{Senmao Li$^{1}$ \quad Joost van de Weijer$^{2}$ \quad Yaxing Wang$^{1}$\thanks{The corresponding author.} \\ \quad Fahad Shahbaz Khan$^{3,4}$ \quad Meiqin Liu$^{5}$ \quad Jian Yang$^{1}$\\
	$^1${VCIP,CS, Nankai University}, $^2${Universitat Aut\`onoma de Barcelona}\\$^3${Mohamed bin Zayed University of AI}, $^4${Linkoping University}, $^5${Beijing Jiaotong University}\\
	{\tt\small  senmaonk@gmail.com \{yaxing,csjyang\}@nankai.edu.cn {joost}@cvc.uab.es}\\ {\tt\small {fahad.khan@liu.se} {mqliu@bjtu.edu.cn}}
}

\maketitle

\begin{strip}
	\vspace{-55pt}
	\centering
	\includegraphics[width=1\linewidth]{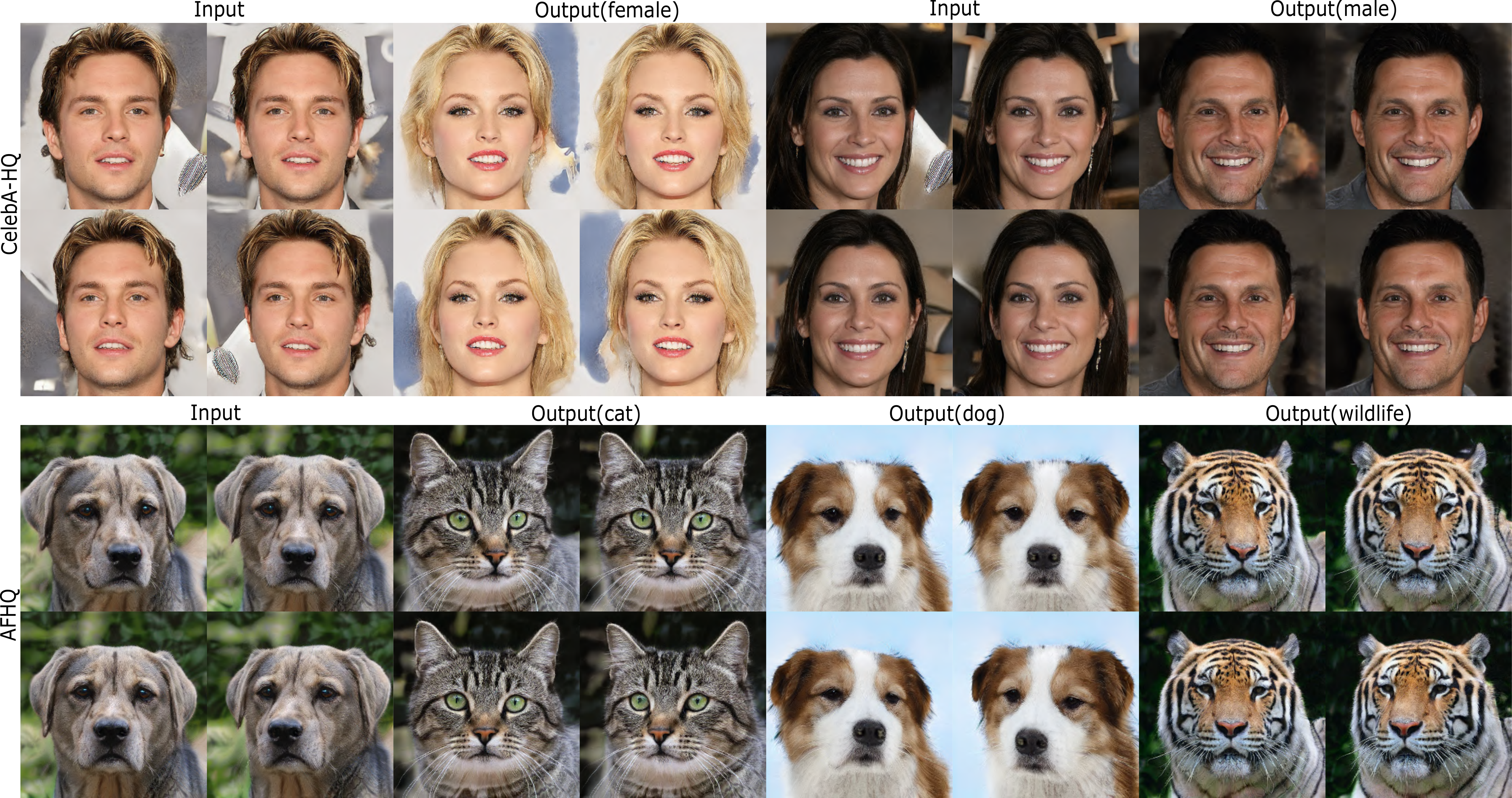}
	\vspace{-19pt}
	\captionsetup{type=figure,font=small}
	\caption{
		3D-aware I2I translation: given a view-consistent 3D scene (the input), our method maps it into a high-quality  target-specific image. Our approach produces consistent results across viewpoints.
	}
        \label{fig:introduction}
	\vspace{-3pt}
\end{strip}

\begin{abstract}
Recent advances in 3D-aware generative models (3D-aware GANs) combined with Neural Radiance Fields (NeRF) have achieved impressive results. However no prior works investigate 3D-aware GANs for 3D consistent multi-class image-to-image (3D-aware I2I) translation. Naively using 2D-I2I translation methods suffers from unrealistic shape/identity change. 
To perform 3D-aware multi-class I2I translation, we decouple this learning process into a multi-class 3D-aware GAN step and a 3D-aware I2I translation step.   In the first step, we propose two  novel techniques: a new conditional architecture and an effective training strategy.  In the second step, based on the well-trained multi-class 3D-aware GAN architecture,
that preserves view-consistency,  
we construct a 3D-aware I2I translation system. To further  
reduce the view-consistency problems, we propose several new techniques, including a U-net-like adaptor network design, a hierarchical representation constrain and a relative regularization loss.   In extensive experiments on two datasets, 
quantitative and qualitative results 
demonstrate  that we successfully perform  3D-aware I2I translation  with  multi-view  consistency.  Code is available in
\href{https://github.com/sen-mao/3di2i-translation}{3DI2I}.
 
\end{abstract}

\section{Introduction}
\label{sec:intro}

 Neural Radiance Fields (NeRF) have increasingly gained attention with their outstanding capacity to synthesize  high-quality view-consistent images~\cite{mildenhall2020nerf,liu2020neural,zhang2020nerf++}. 
Benefiting from the adversarial mechanism~\cite{goodfellow2014generative}, StyleNeRF~\cite{gu2021stylenerf} and concurrent works~\cite{or2022stylesdf,chan2022efficient,deng2022gram,zhou2021cips} have successfully synthesized high-quality view-consistent, detailed 3D scenes by combining NeRF with StyleGAN-like generator design~\cite{karras2019style}. 
This recent progress in 3D-aware image synthesis has not yet been extended to 3D-aware I2I translation, where the aim is to translate in a 3D-consistent manner from a source scene to a target scene of another class (see Figure~\ref{fig:introduction}).

A naive strategy is to use well-designed 2D-I2I translation methods~\cite{isola2016image,park2020contrastive,zhu2017unpaired,huang2018multimodal,Lee2018drit,yu2019multi,Ko_2022_CVPR,yang2022unsupervised}. These methods, however, suffer from
unrealistic shape/identity changes when changing the viewpoint, which are especially notable when looking at a video. Main target class characteristics, such as hairs, ears, and noses, are not geometrically realistic, leading to unrealistic results which are especially disturbing when applying I2I to translate videos.
Also, these methods typically underestimate the viewpoint change and result in target videos with less viewpoint change than the source video.
Another direction is to apply video-to-video synthesis methods~\cite{wang2018video,bansal2018recycle,bashkirova2018unsupervised,chen2019mocycle,liu2021unsupervised}. These approaches, however,  either  rely heavily on labeled data or multi-view frames for each object. In this work, we assume that we only have access to single-view RGB data.

To perform 3D-aware I2I translation, we extend the theory developed for 2D-I2I with recent developments in 3D-aware image synthesis. We decouple the learning process into a multi-class 3D-aware generative model step and a 3D-aware I2I translation step. 
The former can synthesize view-consistent 3D scenes given a scene label, thereby addressing the 3D inconsistency problems we discussed for 2D-I2I. We will use this 3D-aware generative model to initialize our 3D-aware I2I model. It therefore inherits the capacity of synthesizing 3D consistent images.
To train effectively a multi-class 3D-aware generative model (see Figure~\ref{fig:framework}(b)), we provide a new training strategy consisting of: (1) training an unconditional 3D-aware generative model (i.e., StyleNeRF) and (2) partially initializing the multi-class 3D-aware generative model (i.e., multi-class StyleNeRF)  with the weights learned from StyleNeRF. In the 3D-aware I2I translation step, we design a 3D-aware I2I translation architecture (Figure~\ref{fig:framework}(f)) adapted from the trained multi-class StyleNeRF network. To be specific, we use the main network of the pretrained discriminator (Figure~\ref{fig:framework}(b)) to initialize the encoder $E$ of the 3D-aware I2I translation model (Figure~\ref{fig:framework}(f)), and correspondingly, the pretrained generator  (Figure~\ref{fig:framework}(b)) to initialize the 3D-aware I2I generator (Figure~\ref{fig:framework}(f)).  This initialization inherits the capacity of being  sensitive to the view information. 

Directly using the constructed 3D-aware I2I translation model (Figure~\ref{fig:framework}(f)), there still exists some view-consistency problem.  This is because  of the lack of multi-view consistency regularization, and the usage of the single-view image. 
Therefore, to address these problems
 we introduce several techniques, including a U-net-like adaptor network design, a hierarchical representation constrain and a relative regularization loss.

{
In sum, our work makes the following \textbf{contributions}:
\begin{itemize}[leftmargin=*]
    \item 
    We are the first to explore 3D-aware multi-class I2I translation, which allows generating 3D consistent videos.

    \item We decouple 3D-aware I2I translation into two steps. First, we propose a multi-class StyleNeRF. To train this multi-class StyleNeRF effectively, we provide a new training strategy. 
    The second step is the proposal of a 3D-aware I2I translation architecture.

    \item To further address the view-inconsistency problem of 3D-aware I2I translation, we propose several techniques: a U-net-like adaptor, a hierarchical representation constraint and a relative regularization loss. 
 
    \item On extensive experiments, we considerably outperform existing 2D-I2I systems with our 3D-aware I2I method when evaluating temporal consistency.
    
\end{itemize}
}

\section{Related Works}
\minisection{Neural Implicit Fields.} Using neural implicit fields to represent 3D scenes has shown unprecedented quality. \cite{Michalkiewicz_2019_ICCV,OccupancyNetworks,park2019deepsdf,Peng2020ConvolutionalON,sitzmann2019scene,niemeyer2019differentiable} use 3D supervision  to predict neural implicit fields.  
Recently,  NeRF has shown powerful performance to neural implicit representations. NeRF and its variants~\cite{mildenhall2020nerf,liu2020neural,zhang2020nerf++} utilize a volume rendering technique for reconstructing a 3D scene as a combination of neural radiance and density fields to synthesize novel views.

\minisection{3D-aware GANs}
Recent approaches~\cite{nguyen2019hologan,chan2021pi,niemeyer2021giraffe,niemeyer2021campari,lunz2020inverse,henderson2020leveraging,gadelha20173d,jimenez2016unsupervised,xue2022giraffe,zhang2022multi,tewari2022disentangled3d}  learn neural implicit representations without 3D or multi-view supervisions.  
Combined with the adversarial loss, these methods typically randomly sample viewpoints,  render photorealistic 2D images, and finally optimize their 3D representations.   StyleNeRF~\cite{gu2021stylenerf} and concurrent works~\cite{or2022stylesdf,chan2022efficient,deng2022gram,zhou2021cips} have successfully synthesized high-quality view-consistent, detailed 3D scenes with StyleGAN-like generator design~\cite{karras2019style}. In this paper, we investigate 3D-aware image-to-image (3D-aware I2I) translation, where the
aim is to translate in a 3D-consistent manner from a source scene to a target scene of another class. We combine transfer learning of GANs~\cite{wang2018transferring,wang2020minegan}.

\minisection{I2I translation.}
I2I translation  with GAN~\cite{isola2016image,wang2018mix,wang2020deepi2i,wang2020semi} has increasingly gained
attention in computer vision. Based on the differences of the I2I translation task,   recent works focus on  paired I2I translation~\cite{gonzalez2018image,isola2016image,zhu2017toward},  unpaired I2I translation~\cite{kim2017learning,liu2017unsupervised,mejjati2018unsupervised,park2020contrastive,yi2017dualgan,zhu2017unpaired,yang2022unsupervised,shao2021spatchgan,baek2021rethinking,jeong2021memory,wang2019sdit,wang2021transferi2i,laria2021hyper},   diverse I2I translation~\cite{kim2017learning,liu2017unsupervised,mejjati2018unsupervised,park2020contrastive,yi2017dualgan,zhu2017unpaired} and scalable  I2I translation~\cite{choi2020stargan,lee2020drit++,yu2019multi}. 

However, none of these approaches addresses the problem of 3D-aware I2I. For the 3D scenes represented by neural implicit fields, directly using these methods suffers from view-inconsistency.

\section{Method}\label{sec:method}

\minisection{Problem setting.} Our goal is to achieve 3D consistent multi-class I2I translation trained on single-view data only. The system is designed to translate a viewpoint-video consisting of multiple images (source domain) into a new, photorealistic viewpoint-video scene of a target class. 
Furthermore, the system should be able to handle \textit{multi-class} target domains.
We decouple our learning into a multi-class 3D-aware generative model step and a multi-class 3D-aware I2I translation step.

\subsection{Multi-class 3D-aware generative model}

\begin{figure*}[t]
    \centering
    \includegraphics[width=\textwidth]{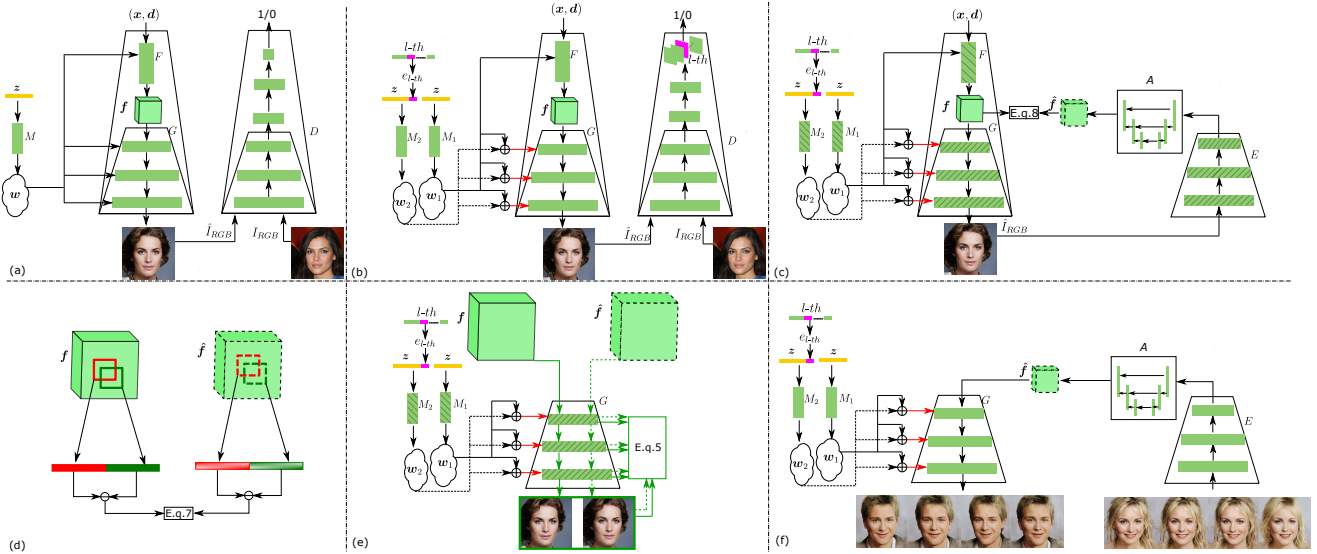}
  \caption{Overview of our method. (a) We first train a 3D-aware generative mode (i.e., StyleNeRF) with single-view photos. (b) We extend StyleNerf to multi-class StyleNerf. We introduce an effective training strategy: initializing  multi-class StyleNeRF with StyleNeRF. (c) The training of the proposed 3D-aware I2I translation. It consists of the encoder $E$, the adaptor $A$,  the generator $G$ and two mapping networks $M_1$ and $M_2$. We freeze all networks except for training the adaptor $A$. The encoder is initialized by the main networks of the pretrained discriminator.  We introduce several techniques to address the view-consistency problems: including a  U-net-like adaptor $A$, (d) relative regularization loss and (e) hierarchical representation constrain. (f) Usage of proposed model at inference time.} \vspace{-5mm}
    \label{fig:framework}
\end{figure*}

Let $\mathcal{I_{RGB}} \in \mathbb{R}^{H\times W\times 3} $ be in the image domain.
In this work, we aim to map a source image  
into a target sample 
conditioned on the target domain label $\vl \in\left\{1,\ldots,L\right\}$ and a random noise vector $\mathbf{z \in \mathbb{R}^{Z}}$.  Let  vector $\vx$ and $\vd$ be 3D location  and 2D viewing direction, respectively.

\minisection{Unconditional 3D-aware generative model.}  
StyleNeRF~\cite{gu2021stylenerf} introduces a 5D function (3D location $\vx$ and 2D viewing direction $\vd$) to predict the volume density $\sigma$ and RGB color $\vc$. Both  $\sigma$ and  $\vc$ are further used to render an image. 
As shown on Figure~\ref{fig:framework}(a)  StyleNeRF consists of four subnetworks: a mapping network $M$, a fully connected layer $F$, a generator $G$ and a discriminator $D$.  The mapping network $M$ takes random noise $\vz$ as input, and outputs latent code $\vw$, which is further fed into both the fully connected layer $F$ and generator $G$.  Given the 3D location $\vx$, the 2D viewing direction $\vd$ and latent code $\vw$,  StyleNeRF  renders the feature map $\vf$: 
\begin{equation}
\begin{aligned}
   & \vf(\bm{r}) = \int_0^\infty\!\! p(t)\vc(\vr(t),\vd) dt\\
   &  p(t) = \exp\parr{ - \int_0^t\!\!  \sigma(\vr(s)) ds }\cdot\sigma_\vw(\vr(t))\\
    & \vc, \sigma = F(\vx, \vd, \vw),
    \label{eq:rendering}
\end{aligned}
\end{equation}

where $\bm{r}(t)=\bm{o} + t\bm{d}$ ($\vo$ is the camera origin) is  a camera ray for each feature representation position. Generator $G$ takes as an input the representation $\vf$  and the latent code $\vw$, and outputs view-consistent photo-realistic novel result $\hat{I}_{RGB}$.  The discriminator $D$ is to distinguish  real images $I_{RGB}$ from generated images $\hat{I}_{RGB}$.  

The fully objective of StyleNeRF is as following:
\begin{equation}
\begin{aligned}\label{eq:uncon_nerf}
    &\mathcal{L}_{G}=\mathbb{E}_{\vz\sim\mathcal{Z},\vp\sim\mathcal{P}}\left[v(D(G(F(\vz,\vx,\vd), M(\vz)))\right] \\
    &+ \mathbb{E}_{I_{RGB}\sim p_{\text{data}}}\left[v(-D(I_{RGB}) + \lambda \|\nabla D(I_{RGB})\|^2)\right]\\
    &+ \beta\cdot \mathcal{L}_{\text{NeRF-path}}
\end{aligned}
\end{equation}
where $v(u)=-\log(1+\exp(-u))$, and $p_{\text{data}}$ is the data distribution. $\mathcal{L}_{\text{NeRF-path}}$ is NeRF path regularization used in StyleNeRF. We also set $\beta=0.2$ and $\lambda=0.5$ following StyleNeRF.

\minisection{Conditional 3D-aware generative model.} 

Figure~\ref{fig:framework}(b) shows the proposed multi-class 3D-aware generative model (i.e., multi-class StyleNeRF).  Compared to the StyleNeRF architecture (Figure~\ref{fig:framework}(a)), we introduce  two mapping networks:  $M_1$ and $M_2$. The mapping network $M_1$ outputs the latent code $\vw_1$. While the mapping network $M_2$  takes as input the concatenated noise $\vz$ and class embedding $e_{\vl\textrm{-}{th}}$, and outputs the latent code $\vw_2$. The second mapping network $M_2$ aims to guide the generator $G$ to synthesize a class-specific image. Here we do not feed the latent code $\vw_2$ into NeRF's fully connected layer $F$, since we expect  $F$ to learn a class-agnostic feature representation, which contributes to perform multi-class 3D-aware I2I translation.

To be able to train multi-class StyleNeRF we adapt the loss function. We require $D$ to address multiple adversarial classification tasks simultaneously, as in~\cite{liu2019few}. 
Specifically, given output $D \in \mathbb{R}^{L}$, we locate the $\vl\textrm{-}{th}$ class response. 

Using the response for the $\vl\textrm{-}{th}$ class, we compute the adversarial loss and back-propagate gradients:  
\begin{equation}
\begin{aligned}
    &\mathcal{L}^{l}_{G}=\mathbb{E}_{\vz\sim\mathcal{Z},\vx\sim\mathcal{P}_{x},\vd\sim\mathcal{P}_{d}}\left[v(D(G(\hat{I}_{RGB}))_{\vl\textrm{-}{th}}\right] \\
    &+ \mathbb{E}_{I_{RGB}\sim p_{\text{data}}}\left[v(-D(I_{RGB})_{\vl\textrm{-}{th}} + \lambda \|\nabla D(I_{RGB})_{l_{th}}\|^2)\right]\\
    &+ \beta\cdot \mathcal{L}_{\text{NeRF-path}}.
\end{aligned}
\end{equation}
We initialize the multi-class StyleNeRF with the weights learned with the unconditional StyleNeRF (E.q.~\ref{eq:uncon_nerf}), since the training from scratch fails to convergence. Results of this are show in Figs.~\ref{fig:multi_class_stylenerf_afhq}. To be specific, we directly copy the weights from the one learned from StyleNeRF for $M_1$, $F$ and $G$ with the same parameter size. For the mapping network $M_2$, we duplicate the weight from $M$ except for the first layer, which is trained from scratch because of the different parameter sizes. The discriminator is similarly initialized except for the last layer, which is a new convolution layer with $L$ output channels. Using the proposed initialization method, we successfully generate class-specific photorealistic high-resolution result.

\subsection{3D-aware I2I translation}

 Figure~\ref{fig:framework} (f) shows the 3D-aware I2I translation network at inference time. It consists of the encoder $E$,  the generator $G$ and two mapping networks $M_1$ and $M_2$. Inspired by DeepI2I~\cite{wang2020deepi2i}, we use the pretrained discriminator (Figure~\ref{fig:framework}(b)) to initialize the encoder $E$ of the 3D-aware I2I translation model (Figure~\ref{fig:framework}(f)), and correspondingly, the pretrained generator (Figure~\ref{fig:framework}(b)) to initialize the 3D-aware I2I generator.   
  To align the encoder with the generator, ~\cite{wang2020deepi2i} introduces a Resnet-like adaptor network  to communicate the encoder and decoder. The adaptor is trained without any real data. However, directly using these techniques for  3D-aware I2I translation still suffers from some view-consistency problems. 
  Therefore, in the following, we introduce several designs to address this problem: a U-net-like adaptor network design, a hierarchical representation constrain and a relative regularization loss.

\minisection{U-net-like adaptor.}  As shown in Figure~\ref{fig:framework}(c), to overcome 3D-inconsistency in the results, we propose a U-net-like adaptor $A$. This design contributes to preserve the spatial structure of the input feature. This has been used before for semantic segmentation tasks and label to image translation~\cite{pix2pix2017}.  In this paper, we experimentally demonstrate that the U-net-like adaptor is effective to reduce the inconsistency. 

\minisection{Hierarchical representation constrain.} 

As shown in Figure~\ref{fig:framework}(e),  given the noise $\vz$, 3D location $\vx$ and 2D viewing direction $\vd$ the fully connected layer $F$ renders the 3D-consistent feature map $\vf = F(\vx,\vd,\vw_1) = F(\vx,\vd,M1(\vz))$.  We further extract the hierarchical representation $ \left \{ G(\vf,\vw_1, \vw_2)_{k} \right \}$ as well as  the synthesized image $\hat{I}_{RGB} = G(\vf,\vw_1, \vw_2)$. 
Here $G(\vf,\vw_1, \vw_2)_{k}$ is the $k \textrm{-}{th} (k = m,\ldots,n , (n>m))$ ResBlock~\footnote{After each ResBlock the feature resolution is half of the previous one in the encoder,  and  two times in generator. In the generator, the last output is image.} output of the generator $G$. We then take the generated image $\hat{I}_{RGB}$ as input for the encoder $E$: $E(\hat{I}_{RGB})$, which is fed into the adaptor network $A$,  that is $\hat{\vf} = A(E(\hat{I}_{RGB}))$. In this step, our loss is

\vspace{-3pt}
\begin{equation}
\begin{aligned}
 \mathcal{L}_{A}= \left \| \vf - \hat{\vf} \right \|_{1}.
\end{aligned}
\end{equation}
For the intermediate layers, we propose a hierarchical representation constrain. Given the output $\hat{\vf}$ and the latent codes (i.e., $\vw_1$ and $\vw_2$)~\footnote{Both $\vw1$ and $\vw2$ are the ones used when generating image $\hat{I}_{RGB}$}, we similarly  collect the hierarchical feature  $\left \{ G(\hat{\vf}, \vw_1, \vw_2)_{k} \right \}$. The objective is 
\vspace{-3pt}
\begin{equation}
\begin{aligned}
 \mathcal{L}_{H}= \sum _{k}  \left \| G(\vf,\vw_1, \vw_2)_{k} -  G(\hat{\vf}, \vw_1, \vw_2)_{k} \right \|_{1}.
\end{aligned}
\end{equation}
In this step, we freeze every network except for the U-net-like adaptor which is learned.  Note that we do not access to any real data to train the adaptor, since we utilize the generated image with from the trained generator (Figure~\ref{fig:framework}(b)).
 
\minisection{Relative regularization loss.} We expect to input the consistency of the translated 3D scene with single-image regularization~\footnote{More precisely, that is the feature map in this paper.} instead of the images from the consecutive views. We propose a relative regularization loss based on neighboring patches. We assume that neighboring patches are equivalent to that on corresponding patches of two consecutive views. For example, when inputting multi-view consistent scene images, the position of eyes are consistently moving.  The fully connected layers (i.e., NeRF mode) $F$ renders the view-consistent feature map $\vf$, which finally decides  the view-consistent reconstructed 3D scene. Thus, we expect the output $\hat{\vf}$ of the adaptor $A$ to obtain the view-consistent property of the feature map $\vf$.  

We randomly sample one vector from the feature map $\vf$ (e.g., {red} square in (Figure~\ref{fig:framework}(d))), denoted as $\vf^{\eta}$.   Then we sample the \emph{eight} nearest neighboring vectors of $\vf^{ \eta }$ ({dark green} square in Figure~\ref{fig:framework}(d))), denoted by $\vf^{\eta,\varepsilon }$ where $\varepsilon =1, \cdots, 8$ is the neighbor index. Similarly, we sample vectors  $\hat{\vf}^{ \eta }$ and $\hat{\vf}^{\eta,\varepsilon }$ from the feature map $\hat{\vf}$ ({red} and {dark green} dash square in Figure~\ref{fig:framework}(d))). We then compute the patch difference:
\begin{equation}
\label{equ6}
\begin{aligned}
    d_{\vf}^{\eta,\varepsilon} = \vf^{\eta} \ominus \vf^{\eta,\varepsilon }&,~ 
    d_{\hat{\vf}}^{\eta,\varepsilon} = \hat{\vf}^{ \eta } \ominus \hat{\vf}^{\eta,\varepsilon },
\end{aligned}    
\end{equation}
where $\ominus$ represents vector subtraction.  In order to preserve the consistency,   we force these patch differences to be small:
\begin{equation}
\label{equ7}
\begin{aligned}
    \mathcal{L}_{R}  = \left \|d_{\vf}^{\eta,\varepsilon} - d_{\hat{\vf}}^{\eta,\varepsilon} \right \|_{1}.
\end{aligned}    
\end{equation}
The underlying intuition is straightforward: the difference vectors of the same location should be most relevant in the latent space compared to other random pairs.

The final objective is 

\vspace{-3pt}
\begin{equation}
\begin{aligned}
 \mathcal{L}= \mathcal{L}_{H} + \mathcal{L}_{A} + \mathcal{L}_{R}.
\end{aligned}
\end{equation}

\begin{table}[t]
    \setlength{\tabcolsep}{1mm}
    \resizebox{\columnwidth}{!}{%
    \centering
    \footnotesize
    \setlength{\tabcolsep}{10pt}
    \begin{tabular}{|c|c|c|c|c|c|}
    \hline
    \multirow{2}{*}{\diagbox{Method}{Dataset}} &\multicolumn{2}{c|}{CelebA-HQ}&\multicolumn{2}{c|}{AFHQ }\cr\cline{2-5}& TC$\downarrow$ 
      & FID$\downarrow$ & TC$\downarrow$& FID$\downarrow$   \cr\cline{2-5}
 \hline
      *MUNIT  &30.240 & 31.4	&28.497&41.5	\cr\cline{1-5}
      *DRIT  &35.452 & 52.1	&25.341&95.6		\cr\cline{1-5}
      *MSGAN  &31.641 & 33.1	&34.236&61.4		\cr\cline{1-5}
      StarGANv2 &10.250 &\textbf{13.6}&3.025 &16.1	 	\cr\cline{1-5} \hline
      Ours  (3D)&\textbf{3.743} & 22.3&\textbf{2.067} &\textbf{15.3}	\cr\cline{1-5}
    \hline \hline
    \multirow{1}{*}{} 
    & TC$\downarrow$ 
      & (unc)FID$\downarrow$ & TC$\downarrow$& (unc)FID$\downarrow$     \cr\cline{2-5}
 \hline
     $\dagger$Liu \textit{et al.}~\cite{liu2021smoothing}   &13.315 & 17.8&3.462 &20.0	\cr\cline{1-5} 
   StarGANv2  &10.250  & 12.2&3.025 &\textbf{9.9}		\cr\cline{1-5}
    $\dagger$Kunhee \textit{et al.}~\cite{kim2022style}  &10.462 & \textbf{6.7}	&3.241 & 10.0		\cr\cline{1-5}
    Ours  (3D)&\textbf{3.743} & 18.7	&\textbf{2.067} &11.4		\cr\cline{1-5}
    \end{tabular}  
    }
\caption{\small Comparison with baselines on TC and FID metrics.* denotes that we used the results provided by StarGANv2. $\dagger$ means that we used the pre-trained networks provided by authors.
}\vspace{-5mm}\label{tab:text_audio_ss_to_image}
\end{table}

\begin{table}[t]
\centering
{\setlength{\tabcolsep}{20pt}\renewcommand{\arraystretch}{1.0}
\resizebox{\columnwidth}{!}{%
\begin{tabular}{cccc|cc}
\hline
Ini.  &Ada.   &Hrc. &Rrl. &TC$\downarrow$   &FID$\downarrow$ 
\\ 
\hline  
 Y  & N  & N  & N   &{2.612}    &{23.8}   \\
 Y  & Y  & N  & N   &{2.324}    &{23.1}  \\
 Y  & Y  & Y  & N   &{2.204}    &{16.1}  \\
 Y  & Y  & Y  & Y   &\textbf{2.067}    &\textbf{15.3}  \\
\hline
\end{tabular}
}
}
\caption{Impact of several components in the performance on AFHQ. The second row is the case where  the 3D-aware I2I translation model is initialized by weights learned from the multi-class StylyNeRF. Then it is trained with a Resnet-based adaptor and $L_1$ loss between the representations $\vf$ and   $\hat{\vf}$.  The proposed techniques continuously improve the consistency and performance.  Ini.: initialization method for multi-class StyleNeRF, Ada.: U-net-like adaptor, Hrc.: Hierarchical representation constrain, Rrl: Relative regularization loss.}
\label{tab:ablation}
\end{table}

\begin{figure}[t]
    \centering
    \includegraphics[width=\columnwidth]{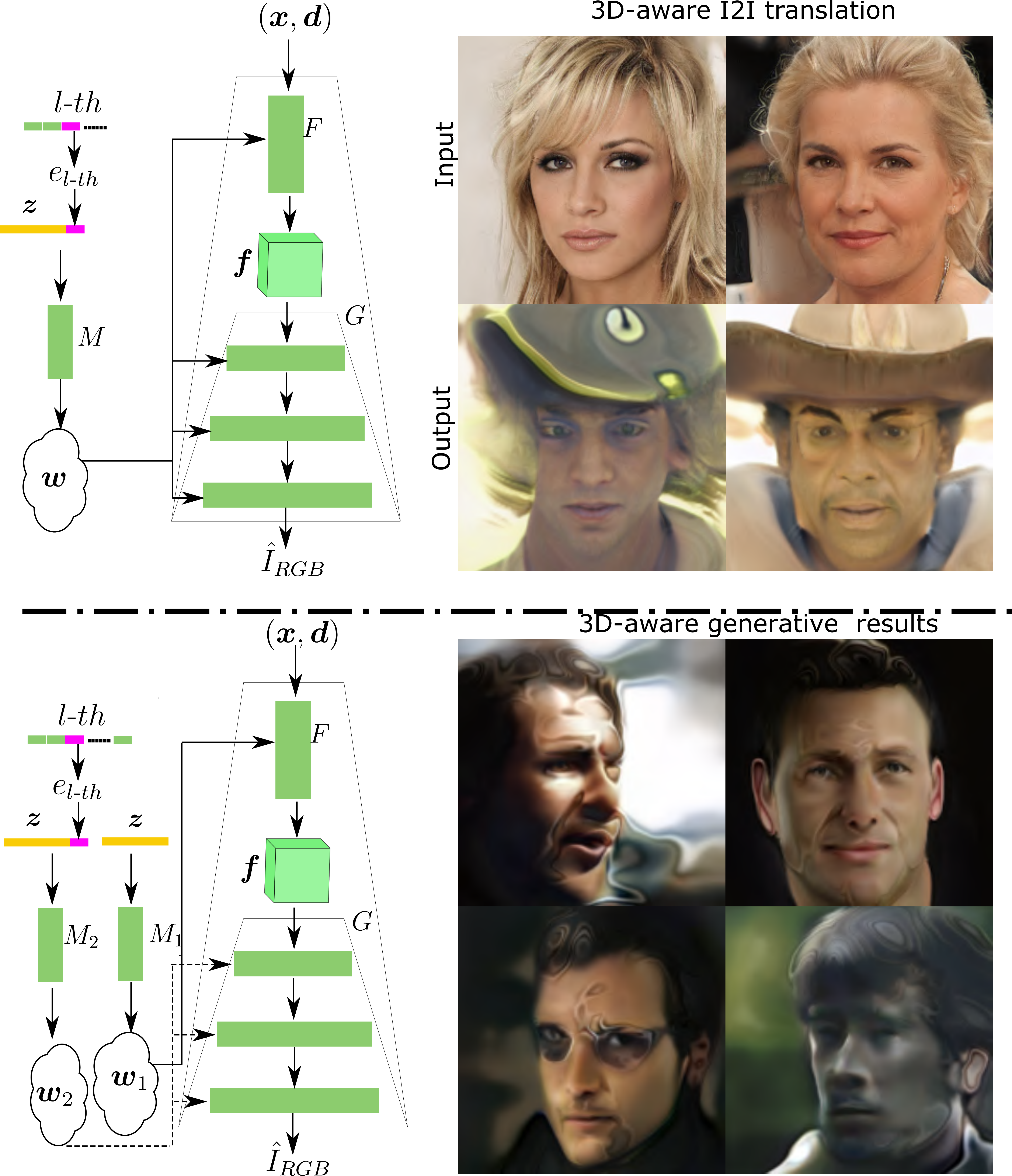}
        \caption{ (Top) Using a single mapping network which takes as input the concatenated class embedding and the noise. We find it fails to generate target-specific realistic image. (Bottom)  we use two mapping networks without concatenating their outputs like the proposed method.   This design fails to generate 3D-aware results.}\vspace{-4mm}
    \label{fig:conditional_network}
\end{figure}

\begin{figure*}[t]
    \centering
    \includegraphics[width=\textwidth]{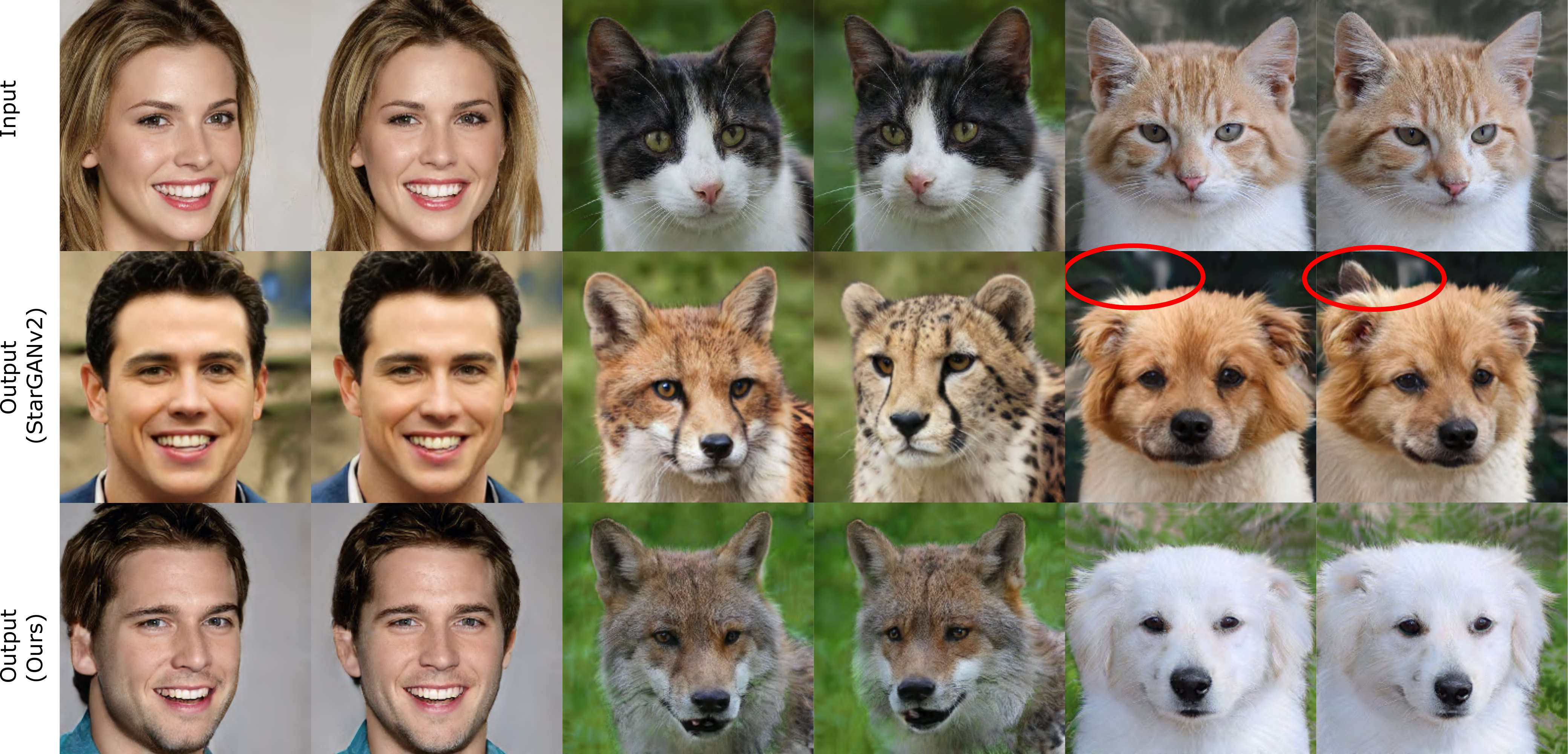}
        \caption{Comparative results between the proposed method and StarGANv2. We observe that StarGANv2 suffers  from underestimating  viewpoint changes when  changing the input viewpoint  ( first column). It also leads to  identity change (third and fourth  columns),   and a geometrically unrealistic ear (last two   columns).}\vspace{-5mm}
    \label{fig:comp_starganv2_ours_afhq}
\end{figure*}

\begin{figure}[t]
    \centering
    \includegraphics[width=\columnwidth]{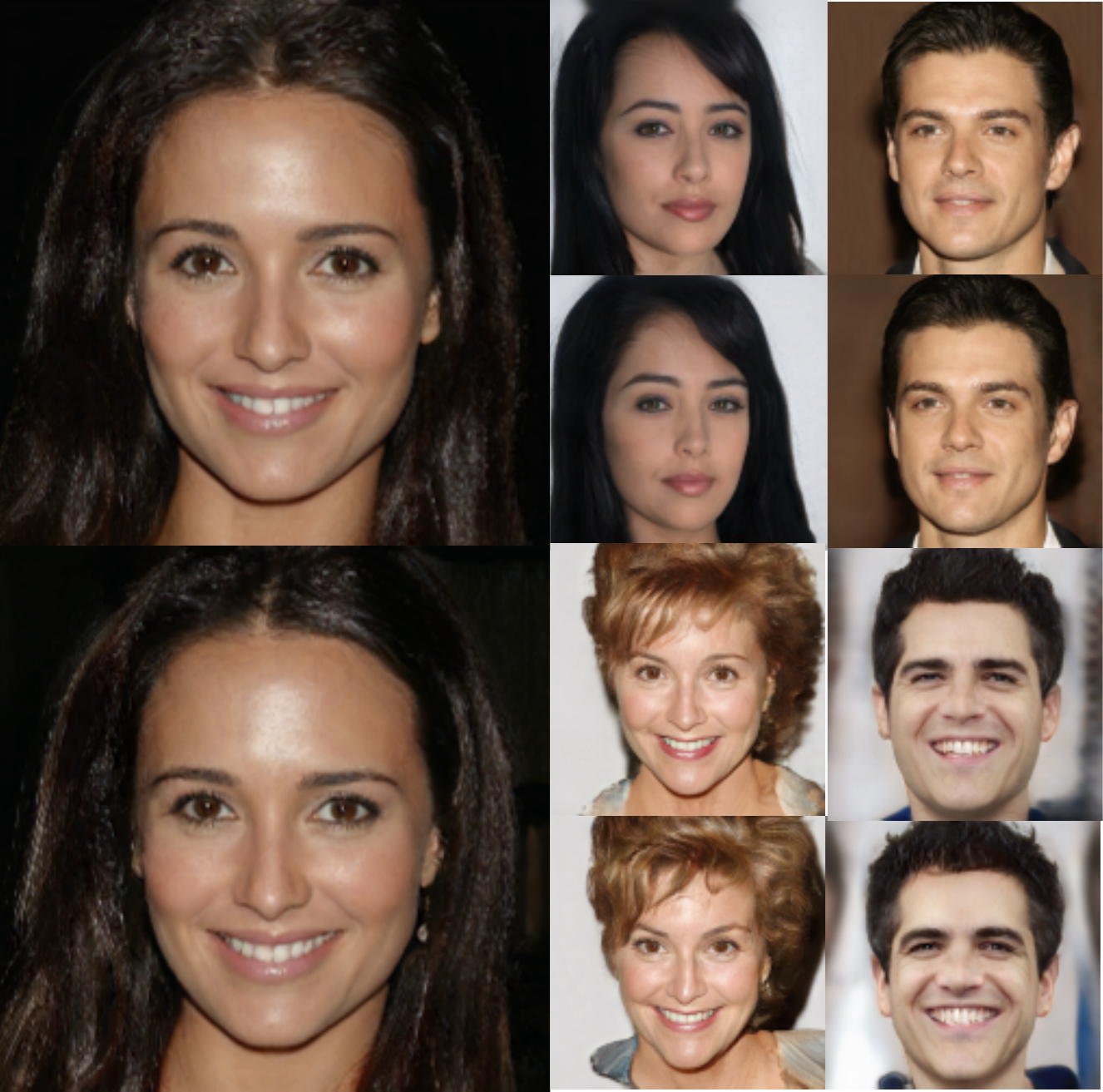}
        \caption{The generated images of (top) $G(\vf, \vw_1, \vw_2)$ and (bottom)  $G(\hat{\vf}, \vw_1, \vw_2)$, which show that we correctly align the outputs of both the NeRF mode  $F$  and the adaptor $A$.}\vspace{-7mm}
    \label{fig:align}
\end{figure}

\section{Experiments}
\label{sec:experiments}

\subsection{Experimental setup}

\minisection{Training details.}
We use the trained StyleNeRF to partially initialize our  multi-class StyleNeRF architecture. We adapt the structure of the multi-class StyleNeRF to the 3D-aware I2I  architecture. The proposed method is implemented in Pytorch~\cite{paszke2017automatic}.  We use  Adam~\cite{kingma2014adam} with a batch size of 64, using a learning rate of 0.0002. We use $2\times$ Quadro RTX 3090 GPUs (24 GB VRAM) to conduct all our experiments. We show the network details and more results on Supp.~Mat..

\minisection{Datasets.} Our experiments are  conducted on the Animal Faces (AFHQ)~\cite{choi2020stargan} and CelebA-HQ~\cite{karras2017progressive} datasets.  AFHQ contains 3 classes, each one has about 5000 images.  In CelebA-HQ, we use gender as a class, with $\sim$10k(10057) male and $\sim$18k(17943) female images in the training set.  In this paper, all images are resized to $256 \times 256$.

\minisection{Baselines.} We compare to MUNIT~\cite{huang2018multimodal}, DRIT~\cite{Lee2018drit}, MSGAN~\cite{karnewar2020msg},  StarGANv2~\cite{choi2020stargan}, \cite{kim2022style} and \cite{liu2021smoothing}, all of which perform image-to-image translation.

\minisection{Evaluation Measures.} We employ the widely used metric for evaluation, namely Fr\'echet Inception Distance (FID)~\cite{heusel2017gans}. 
 We also propose a new measure in which we combine two metrics, one which measures the consistency between neighboring frames (which we want to be low), and another that measures the diversity over the whole video (which we would like to be high). We adopt a modified \textit{temporal loss} (TL)~\cite{wang2020consistent}. This temporal loss computes the Frobenius difference  between two frames to evaluate the video consistency. Only considering this measure would lead to high scores when neighboring frames in the generated video are all the same. For successful 3D-aware I2I translation, we expect the system to be sensitive to view changes in the source video and therefore combine low consecutive frame changes with high diversity over the video. Therefore, we propose to compute LPIPS~\cite{zhang2018unreasonable} for each video (vLPIPS), which indicates the diversity of the generated video sequence. To evaluate both the consistency and the sensitiveness of the generated video, we propose a new temporal consistency metric (TC):

\begin{equation}
\begin{aligned}
    TC= TL/vLPIPS.
\end{aligned}    
\end{equation}
Due to the small changes between two consecutive  views, for each video we use frame interval 1, 2 and 4  in between to evaluate view-consistency. Note that a lower TC value is better.

\begin{figure}[t]
    \centering
    \includegraphics[width=\columnwidth]{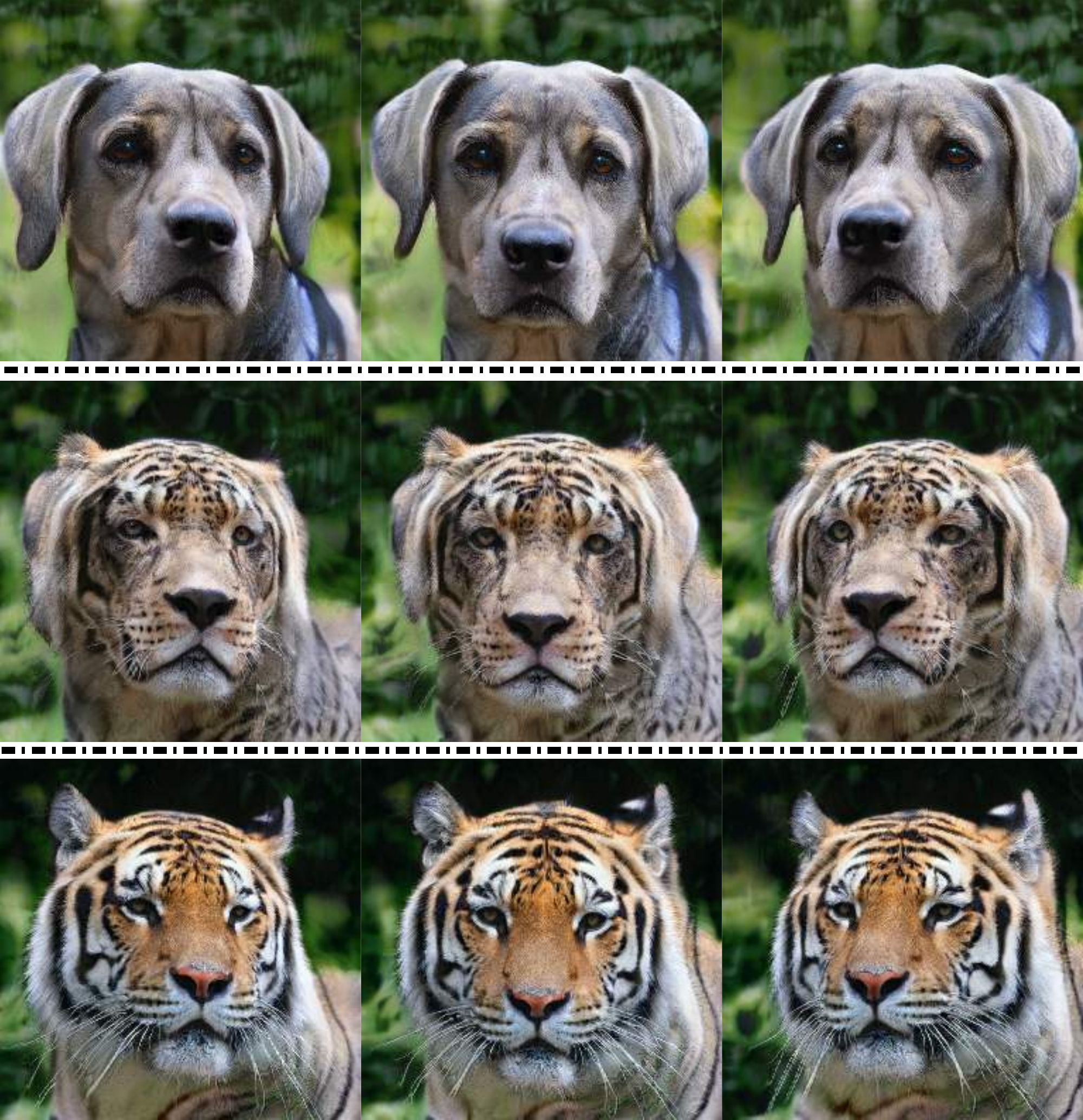}
        \caption{Interpolation between the dog and wildlife classes.}\vspace{-7mm}
    \label{fig:interpolation}
\end{figure}

\begin{figure*}[t]
    \centering
    \includegraphics[width=\textwidth]{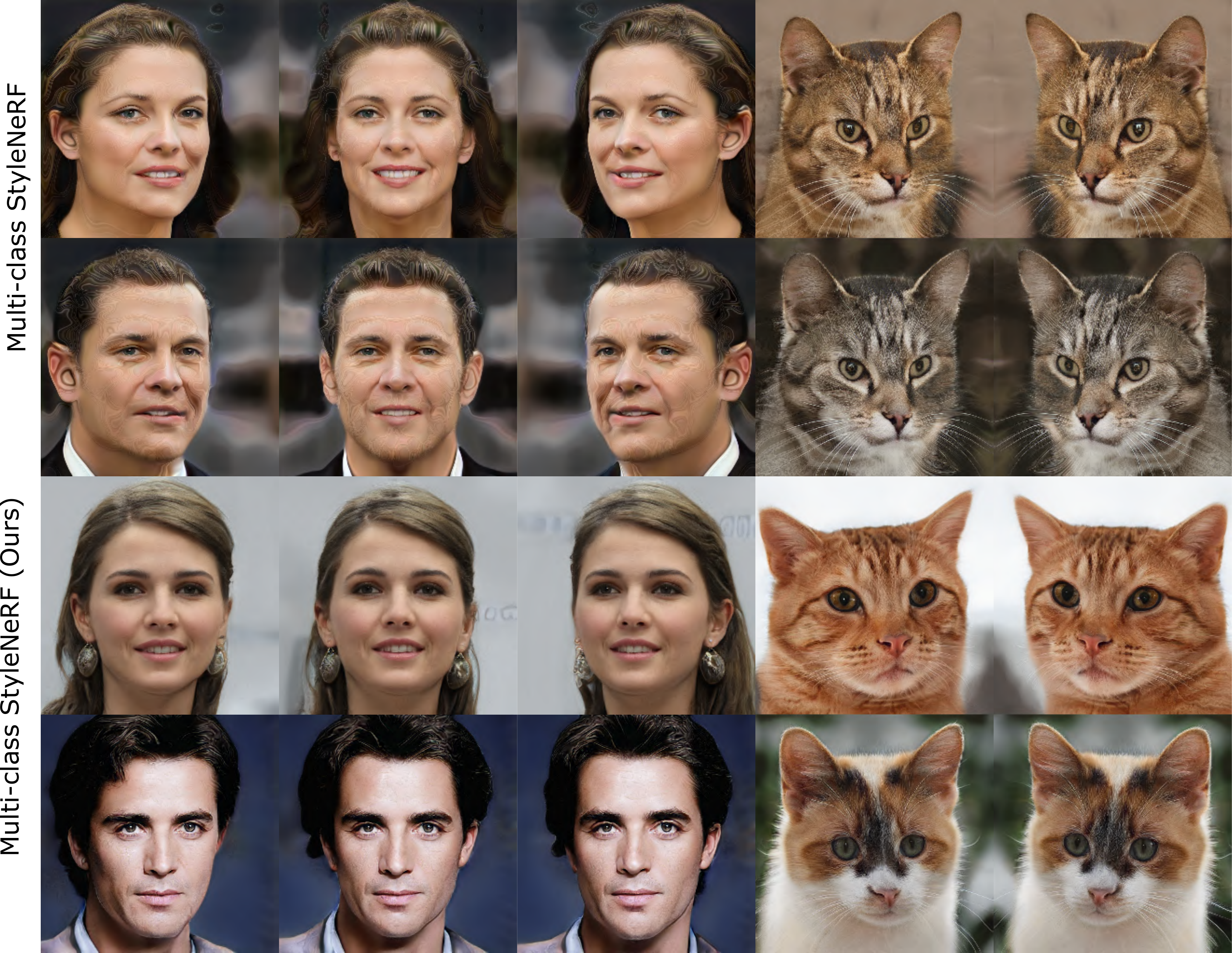}
        \caption{Qualitative results of multi-class StyleNeRF training from scratch (up) and from the proposed strategy (bottom).}\vspace{-4mm}
    \label{fig:multi_class_stylenerf_afhq}
\end{figure*}

\subsection{Quantitative  and qualitative results.}

We evaluate the performance of the proposed method on both the AFHQ animal and CelebA human face dataset. As reported in Table~\ref{tab:text_audio_ss_to_image},  in terms of TC the proposed method achieves the best score on two datasets. For example, we have 3.743 TC on CelebA-HQ, which is better than StarGANv2 (10.250 TC). This indicates that our method dramatically improves  consistency.  As reported in Table~\ref{tab:text_audio_ss_to_image} (up), across both datasets, the proposed method consistently outperforms the baselines with significant gains in terms of FID and LPIPS, except for StarGANv2 which obtains superior results. 
However, on AFHQ we achieve better FID score than StarGANv2. Kunhee \textit{et al.}~\cite{kim2022style} reports the unconditional FID (\textit{(unc)FID}) value which is computed between  synthesized images and training samples instead of each class. As reported in Table~\ref{tab:text_audio_ss_to_image} (bottom),  We are able to achieve completing results on uncFID metrics. 
Note that while 2D I2I translation (e.g., StarGANv2) can obtain high-quality for each image, they cannot synthesize images of the same scene with 3D consistency, and suffers from unrealistic shape/identity changes when changing the viewpoint, which are especially notable when looking at a video.

In Figures~\ref{fig:introduction},\ref{fig:comp_starganv2_ours_afhq},  we perform 3D-aware I2I translation. When  changing the input viewpoint (Figure~\ref{fig:comp_starganv2_ours_afhq} (first two columns)), the outputs of StarGANv2 
do not maintain the correct head pose, and underestimate the pose changes with respect to the frontal view. To estimate that this is actually the case, we also compute the diversity (i.e., vLPIPS) in a single video sequence. For example, both StarGANv2 and our method are 0.032 and 0.101 on CelebA-HQ. This confirms that the diversity (due to pose changes) is lowest for StarGANv2. More clearly showing the limitations of standard I2I methods for 3D-aware I2I, we observe that StarGANv2 suffers from unrealistic changes when changing the viewpoint. For example, when translating the class \emph{cat} to \emph{wildlife}, the generated images changes from \emph{wolf} to \emph{leopard} when varying the viewpoint (Figure~\ref{fig:comp_starganv2_ours_afhq} (third and fourth  columns)). . 
Also, the main target class characteristics, such as ears, are not geometrically realistic, leading to unrealistic 3D scene videos.   Our method, however, 
eliminates these shortcomings  and  performs  efficient high-resolution image translation with high 3D-consistency, which preserves the input image pose and changes the style of the output images. We show high-resolution images ($1024 \times 1024$) on Supp.~Mat..

\subsection{Ablation study}

\minisection{Conditional 3D-aware generative architecture}
In this experiment, we verify our network design by comparing it with two alternative network designs.
As shown in Figure~\ref{fig:conditional_network}(up), we  explore a naive strategy: using one mapping which takes as input the concatenated class embedding and the noise.  In this way, the fully connected network $F$ outputs the \textit{class-specific} latent code $\vw$,  which is fed into the fully connected network $F$ to output the\textit{class-specific} representation $\vf$.   Here, both the latent code $\vw$ and the representation $\vf$ are decided by the same class. However, when handling 3D-aware multi-class I2I translation task,  the feature representation $\hat{\vf}$ is combined with the latent code $\vw$ from varying class embeddings, which leads to unrealistic image generation (Figure.~\ref{fig:conditional_network}(up)). 

As shown in Figure~\ref{fig:conditional_network}(bottom), we utilize two mapping networks without concatenating their outputs like the proposed method.  This design guarantees that the output of the fully connected layers $F$ are \textit{class-agnostic}.  We experimentally observe that this model fails to handle 3D-aware generation.

\minisection{Effective training strategy for multi-class 3D-aware generative model.} We evaluate the proposed training strategy on AFHQ and CelebA-HQ datasets. We initialize  the proposed multi-class 3D I2I architecture  from scratch and the proposed method, respectively. As shown on Figure~\ref{fig:multi_class_stylenerf_afhq} (up),  the model trained from scratch  synthesizes unrealistic faces on CelebA-HQ dataset, and low quality cats on AFHQ. This is due to the style-based conditional generator which is hard to be optimized and causes mode collapse directly~\cite{sauer2022stylegan}. The proposed training strategy, however, manages to synthesize  photo-realistic high-resolution images with high multi-view consistency.  This training strategy first performs unconditional learning, which leads to satisfactory generative ability. Thus, we relax the difficulty of directly training the conditional model. 

\minisection{Alignment and interpolation.}  Figure~\ref{fig:align} exhibits the outputs of the generator when taking as input the feature representation  $\vf$ and  $\hat{\vf}$. This confirms that the proposed method successfully aligns the outputs of the fully connected layers $F$ and the adaptor $A$. Figure~\ref{fig:interpolation} reports interpolation by freezing the input images while interpolating the class embedding
between two classes. Our model still manages to preserve the view-consistency, and generate high quantity images with  even given never seen class embeddings.

\minisection{Techniques for improving the view-consistency.}
We  perform an ablation study on the impact of several design elements on the overall performance of the system, which includes the proposed initialization 3D-aware I2I translation model (Ini.), U-net-like adaptor (Ada.), hierarchical representation constrain (Hrc.) and relative regularization loss (Rrl.). 
We evaluate these four factors in Table~\ref{tab:ablation}.   The results
show that only using the proposed initialization (the second row of the Table~\ref{tab:ablation}) has already improved the view-consistency comparing to StarGANv2 (Table~\ref{tab:text_audio_ss_to_image}).  Utilizing either U-net-like adaptor (Ada.) or hierarchical representation constrain (Hrc.) further leads to performance gains.  Finally we are able to get the best score when further adding relative regularization loss (Rrl.) to the 3D-aware I2I translation model.
\section{Conclusion}
In this paper we first explore 3D-aware I2I translation.   We decouple the learning process into a multi-class 3D-aware generative model step and a 3D-aware I2I translation step.  In the first step, we propose a new multi-class StyleNeRF architecture, and an effective training strategy.  We design the 3D-aware I2I translation model with the well-optimized multi-class StyleNeRF model. It inherits the capacity of synthesizing 3D consistent images. In the second step, we propose several techniques to further reduce  the  view-consistency of the 3D-aware I2I translation. 

\textbf{Acknowledgement.} 
We acknowledge the support from the Key Laboratory of Advanced Information Science and Network Technology of Beijing (XDXX2202), and the project supported by Youth Foundation (62202243). We acknowledge the Spanish Government funding for projects PID2019-104174GB-I00, TED2021-132513B-I00.

{\small
\bibliography{shortstrings,egbib}
\bibliographystyle{ieee_fullname}
}

\clearpage
\appendix
\noindent{\Large\bf Supplementary Material}
\vspace{5pt}

\section{Network detail}
\subsection{U-net-like adaptor}
Inspired by Pix2pix~\cite{pix2pix2017}, we design a U-net-like adaptor between the encoder and the decoder. 
Encoder:
C64-C128-C256-C512-C512
Decoder:
C512-C256-C128-C64-C64
To specific, we apply Batch-norm  and leaky-ReLUs (0.2) after the convolution in the encoder, and  Batch-norm and ReLU after the convolution in the decoder except for the last layer. The last layer of the decoder is a convolution of which the number of output channels is 64.  The U-Net architecture is identical except for the skip connection  between each layer $k$ in the encoder and the layer $n-k$ in the decoder, where $n$ is the number of the decoder.

\subsection{Test detail}
At inference time, like StarGANv2, we use 50000 images to compute FID. To evaluate the view-consistency, we use 100 videos to compute \textit{TC}, and report the mean value.

\section{Ablation}
\subsection{Alignment with both the 3D location and 2D viewing direction}
\begin{figure}[t]
    \centering
    \includegraphics[width=\columnwidth]{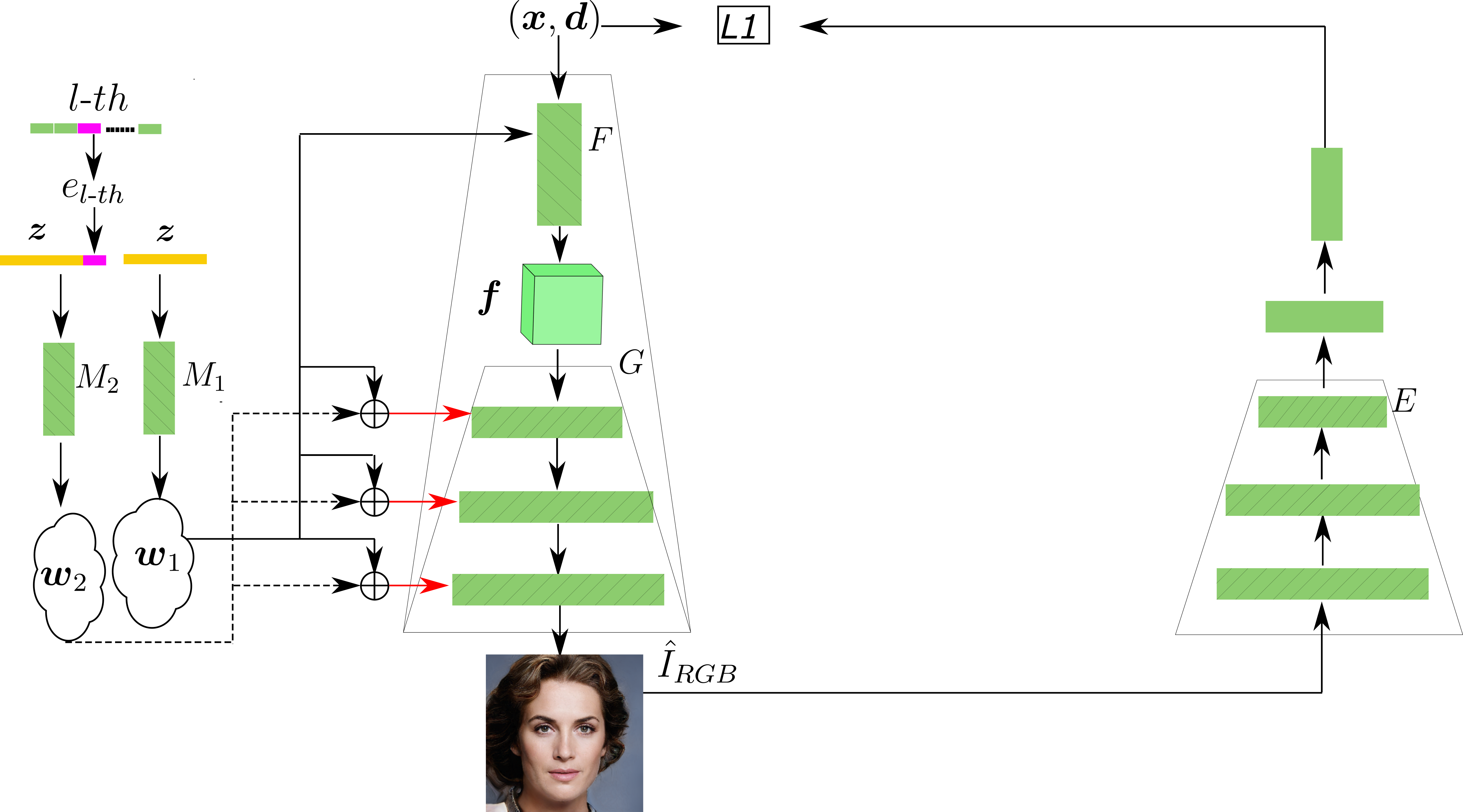}
        \caption{We predict 3D location and 2D viewing direction, and align with the input ones.}
    \label{fig:predict_pose}
\end{figure}

\begin{figure}[t]
    \centering
    \includegraphics[width=\columnwidth]{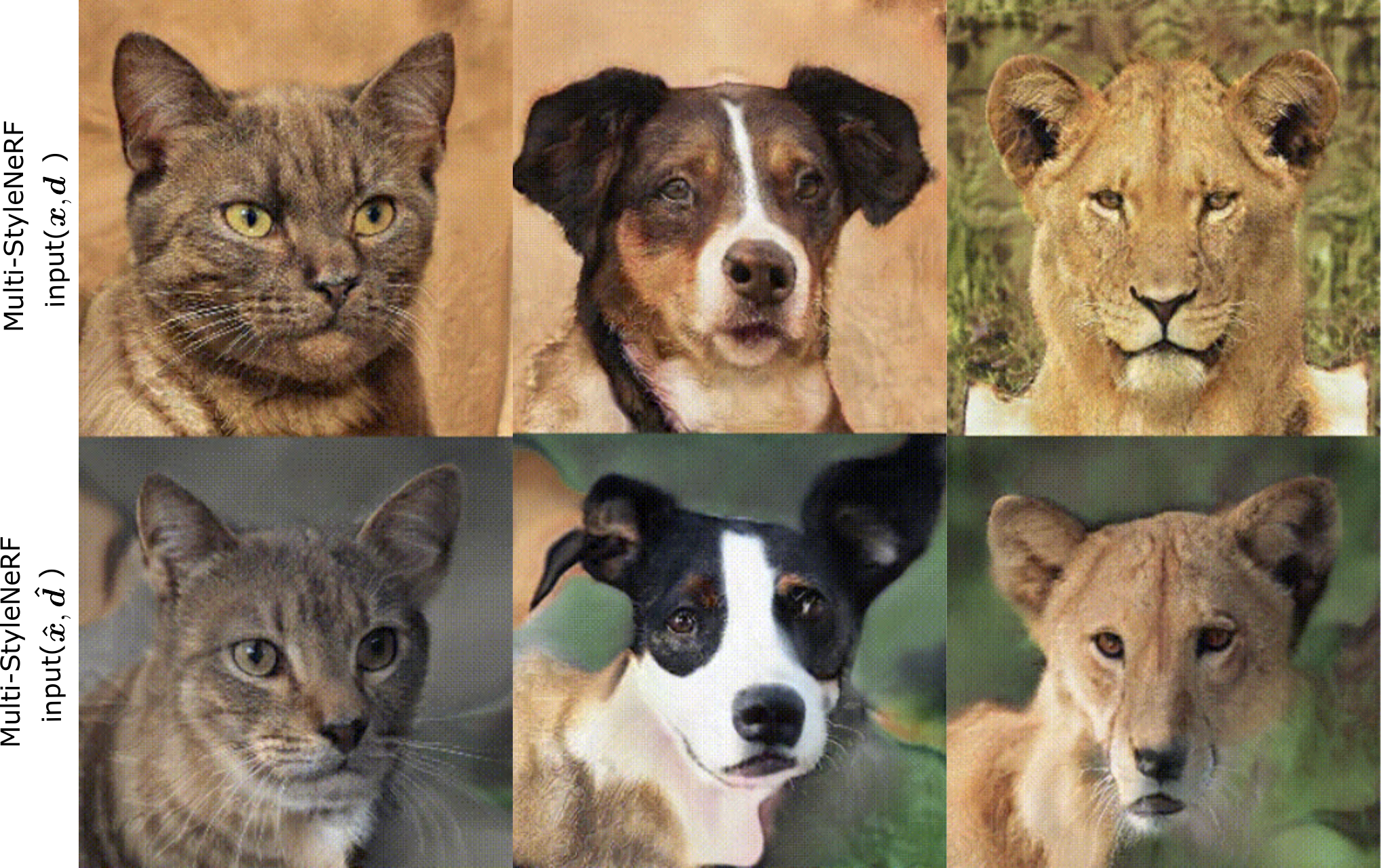}
        \caption{The synthesised images with the input $(\vx, \vd)$ and the predicted  $(\hat{\vx}, \hat{\vd})$.}
    \label{fig:predicting_pose_view}
\end{figure}

Instead of outputting  the feature map  $\hat{\vf}$, we use the adaptor to output the 3D location $\hat{\vx}$ and 2D viewing direction $\hat{\vd}$, and align with the input  the 3D location $\vx$ and 2D viewing direction $\vd$ (Figure~\ref{fig:predict_pose} ).  As shown in Figure~\ref{fig:predicting_pose_view}, when utilizing the predicted   3D location $\hat{\vx}$ and 2D viewing direction $\hat{\vd}$, we are not able to generate high-quality images like the ones synthesized by the location $\vx$ and 2D viewing direction $\vd$.

\section{Unconditional 3D-aware I2I translation}

We also explore the unconditional  3D-aware I2I translation. As shown in Figure~\ref{fig:unconditional_framework} (left), we design  the unconditional  3D-aware I2I translation architecture. To be specific, we share the NeRF$F$, and device the domain-specific generators: $G_1$ and $G_2$.  Figure~\ref{fig:unconditional_framework} (right) shows we test our model. Note, we train this system like multi-class 3D-aware I2I translation.   Figure~\ref{fig:unconditional_results} shows the generated images, which indicates the effectiveness of the proposed method.

\section{Additional results}

\subsection{Quantitative results}

Table.~\ref{tab:vlpips_tl} reports the result of the additional metrics: vLPIPS and TL.  Although StarGANv2 has better performance,  it fails to preserve 3D Consistent. We have best score on \textit{TC}, which is corresponding to consistency.  
\subsection{User study}

\begin{figure}[t]
    \centering
    \includegraphics[width=\columnwidth]{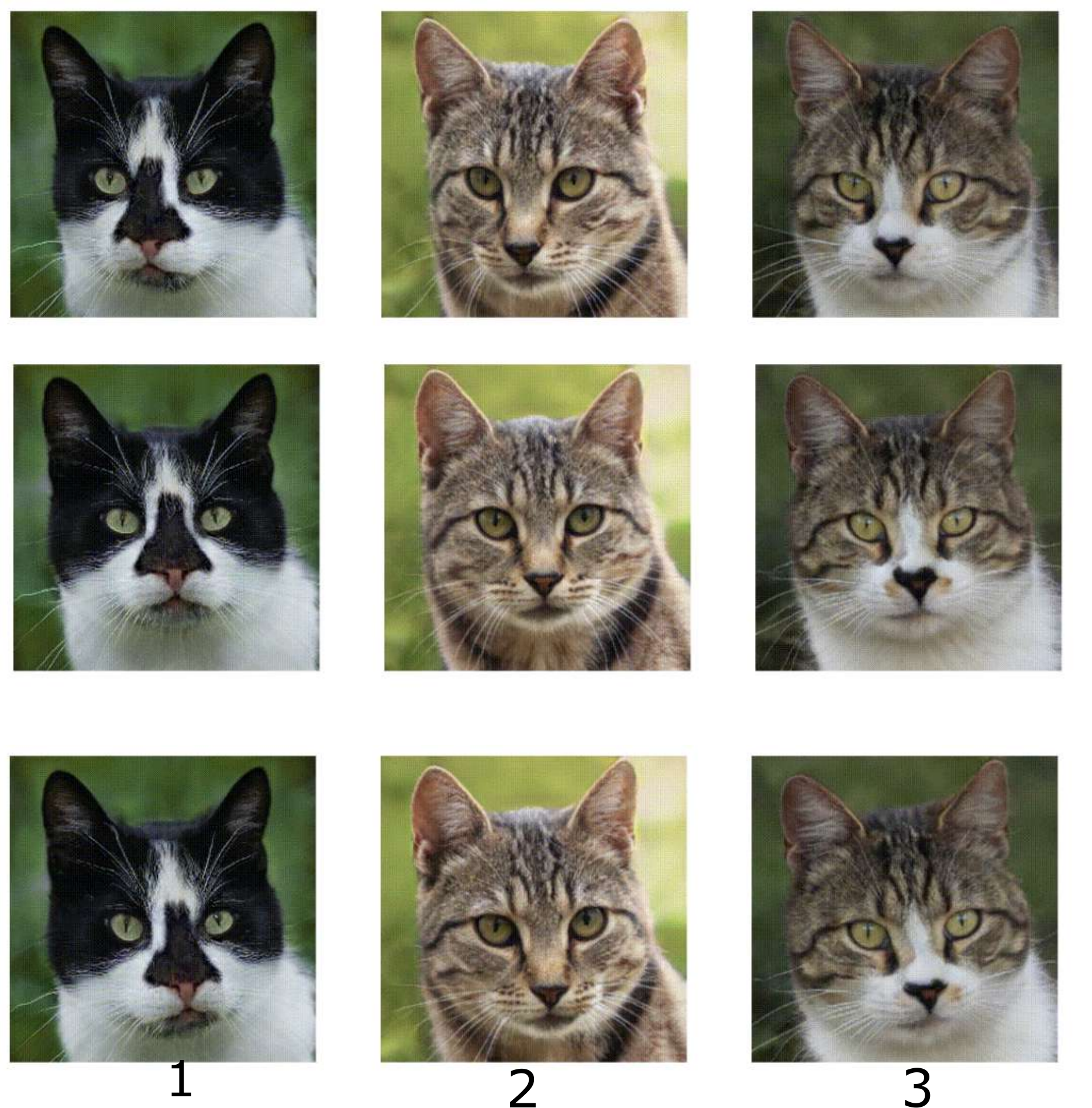}
        \caption{User study example. We conduct a user study and ask subjects to select the results that is \textit{Which video has the best view-consistency? please select one}.}
    \label{fig:user_study_example}
\end{figure}

\begin{figure}[t]
    \centering
    \includegraphics[width=\columnwidth]{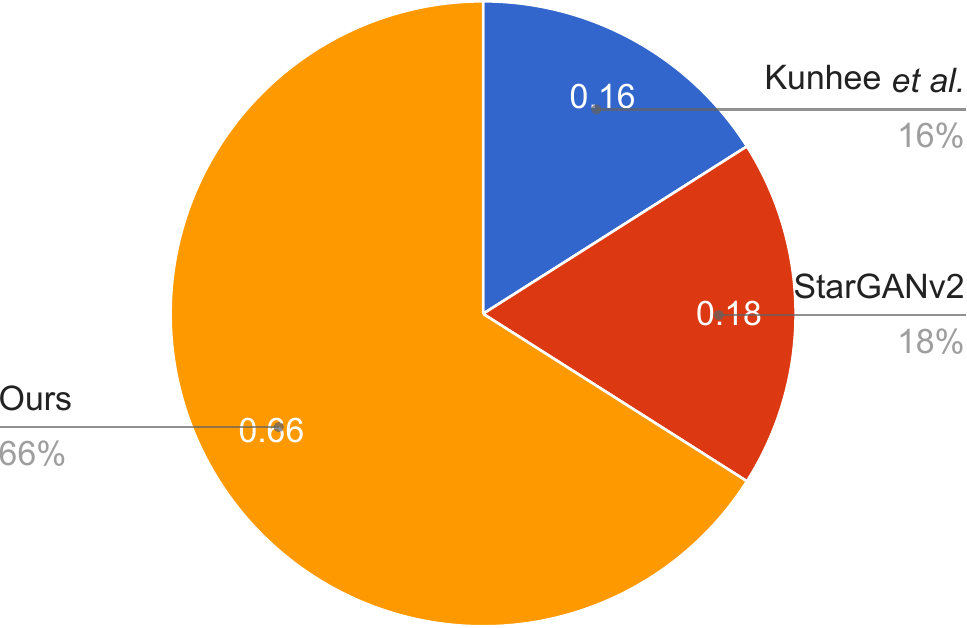}
        \caption{User study.}
    \label{fig:user_study}
\end{figure}

\begin{figure*}[t]
    \centering
    \includegraphics[width=\textwidth]{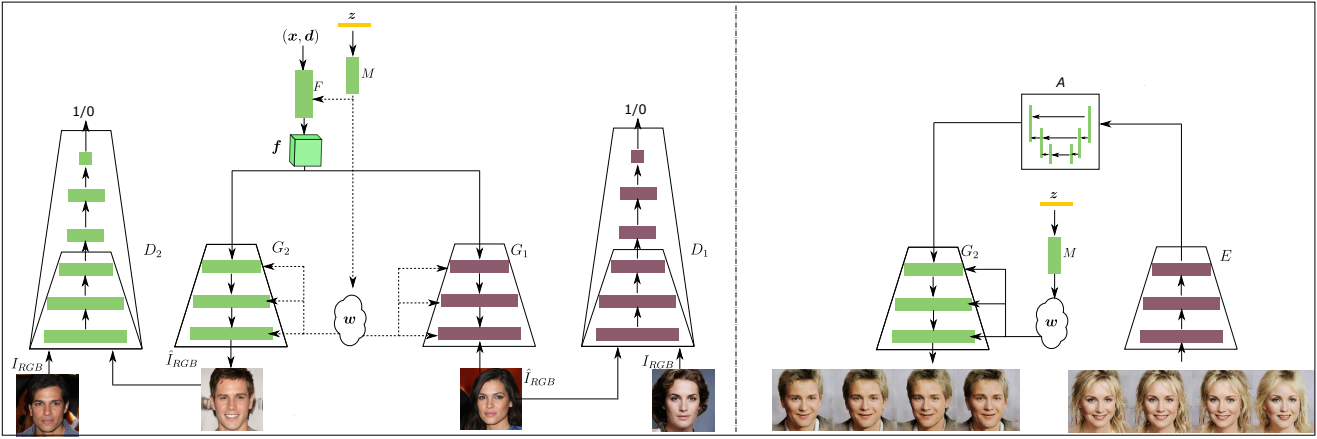}
        \caption{The unconditional 3D-aware I2I translation. (Left) we design unconditonal 3D-aware generative model. Here we share the NeRF mode $F$. (Right) Usage of proposed model at inference time.  Note we train this system like multi-class 3D-aware I2I translation. }
    \label{fig:unconditional_framework}
\end{figure*}

\begin{figure*}[t]
    \centering
    \includegraphics[width=\textwidth]{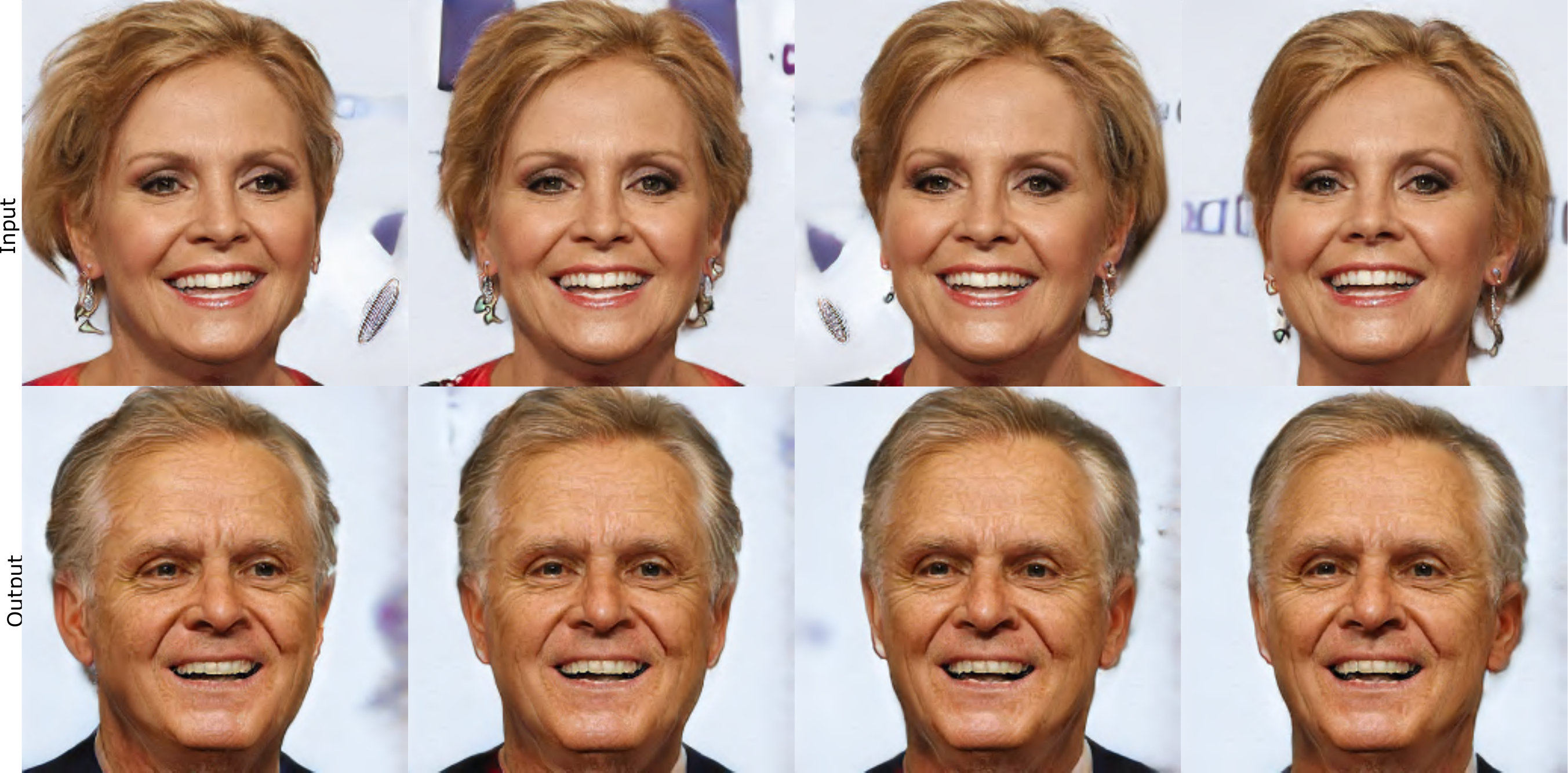}
        \caption{The synthesized images by the proposed unconditional 3D-aware I2I translation.}
    \label{fig:unconditional_results}
\end{figure*}

We conduct a user study and ask subjects to select the results that is \textit{Which video has the best view-consistency? please select one} (Figure~\ref{fig:user_study_example}). We apply triplet comparisons (forced choice) with 14 users (10 triplets/user) for 3D-aware I2I translation. Experiments are performed on images from the AFHQ dataset. Fig.~\ref{fig:user_study} shows that our method considerably outperforms the other methods.

\begin{table}[t]
    \setlength{\tabcolsep}{1mm}
    \resizebox{\columnwidth}{!}{%
    \centering
    \footnotesize
    \setlength{\tabcolsep}{10pt}
    \begin{tabular}{|c|c|c|c|c|c|c|c|c|c|}
    \hline
    \multirow{2}{*}{\diagbox{Method}{Dataset}} &\multicolumn{3}{c|}{CelebA-HQ}&\multicolumn{3}{c|}{AFHQ }\cr\cline{2-7}& TC$\downarrow$ 
      & vLPIPS$\uparrow$ & TL$\downarrow$ & TC$\downarrow$& vLPIPS$\uparrow$ & TL$\downarrow$  \cr\cline{2-7}
    \hline
      StarGANv2 &10.250 &\textbf{0.032}&\textbf{0.328} &3.025	&0.121 &0.366	\cr\cline{1-7} 
      GP-UNIT  &3.065 & 0.123	&0.377&2.073 &\textbf{0.193}	&0.4	\cr\cline{1-7}
    \hline
      Ours  (3D)&\textbf{3.743} & 0.101&{0.378} &\textbf{2.067} & 0.165 &\textbf{0.341}	\cr\cline{1-7}
    \end{tabular}
    }%
    \caption{\small Comparison with baselines on vLPIPS, the temporal loss (TL) and the temporal consistency (TC), where TC=TL/vLPIPS.}
    \vspace{-5mm}
    \label{tab:vlpips_tl}
\end{table}

\subsection{T-SNE}
\begin{figure*}[t]
    \centering
    \includegraphics[width=\textwidth]{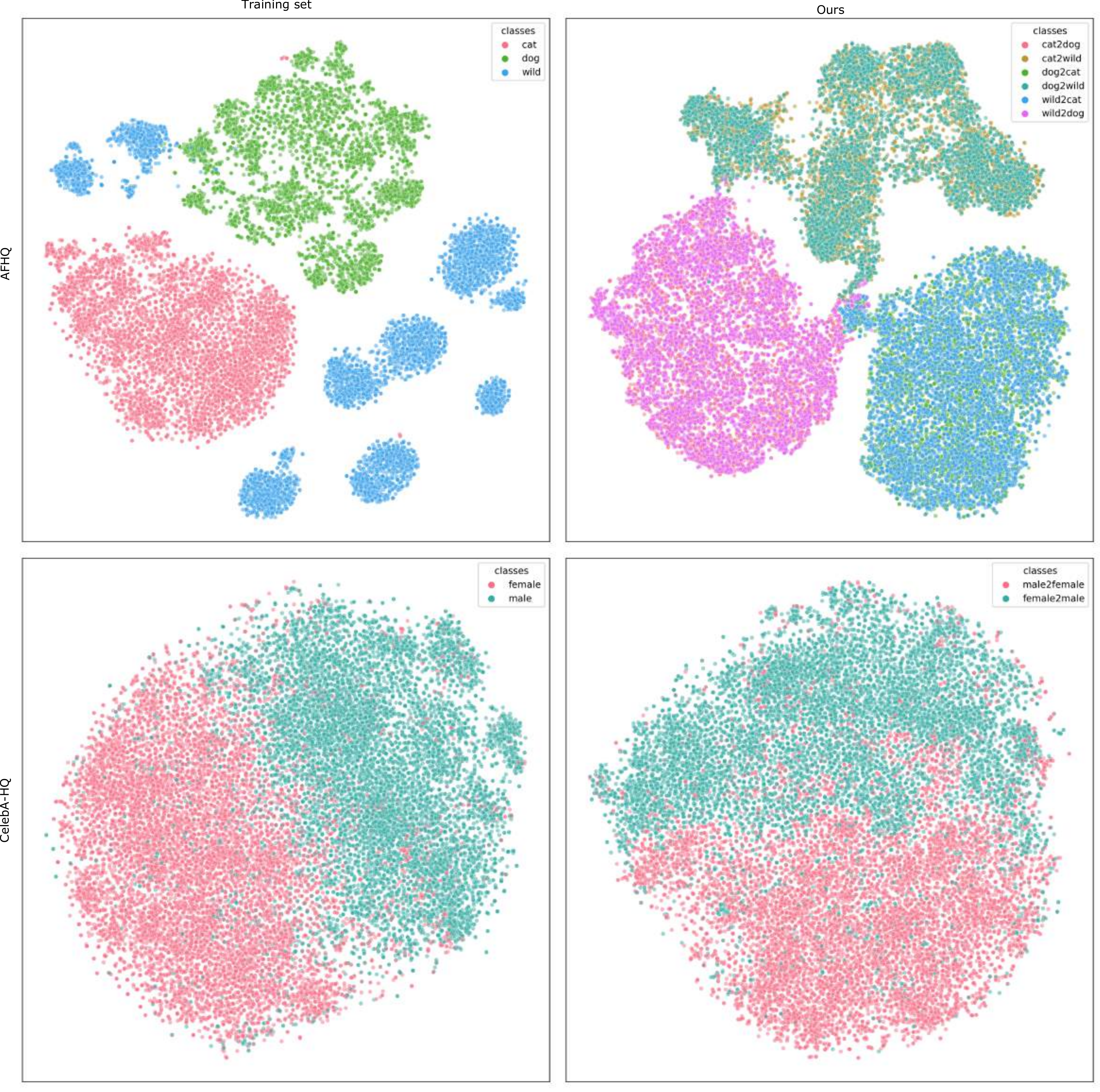}
        \caption{We show T-SNE of both the training sett and the proposed method on two datasets.  We exhibits the translation for each category (top right).  We observe that there is the similar distribution when the target domain is same. }
    \label{fig:tsne}
\end{figure*}

We investigate the latent space of the generated images. We randomly sample images, and translate them into the target class. Specifically, given the generated images we firstly perform Principal Component Analysis (PCA)~\cite{bro2014principal} to the extracted feature. Then, we conduct the T-SNE~\cite{maaten2008visualizing} to visualize the generated images in a two-dimensional space. As illustrated in Figure~\ref{fig:tsne}, 
the T-SNE plot shows that our method has a similar distribution as the training set.

\subsection{Additional Qualitative Results}

We provide additional 3D-aware mult-class I2I translation  interpolation (Figure~\ref{fig:inter1}) and  results for models trained on AFHQ and CelebA-HQ datasets in Figures~\ref{fig:add_afhq2},~\ref{fig:add_afhq3},~\ref{fig:add_afhq4},~\ref{fig:add_afhq5},~\ref{fig:add_afhq6},~\ref{fig:add_celeba_hq1},~\ref{fig:add_celeba_hq2},~\ref{fig:add_celeba_hq3} ,~\ref{fig:add_celeba_hq4} ,~\ref{fig:celeba1024_3},~\ref{fig:celeba1024_4}. 

We also provide the demo for StarGANv2, Kunhee \textit{et al.}~\cite{kim2022style} and ours. While Kunhee \textit{et al.}~\cite{kim2022style}  obtains the best FID score on CelelbA-HQ dataset, it suffers from the view-consistency problem. Please see the accompanying video for more results.

\twocolumn[{
\renewcommand\twocolumn[1][]{#1}
\maketitle
\vspace{-0.7 cm}
\begin{center}
    \centering 
    \vspace{-0.0 cm}
        \includegraphics[width=0.88\textwidth]{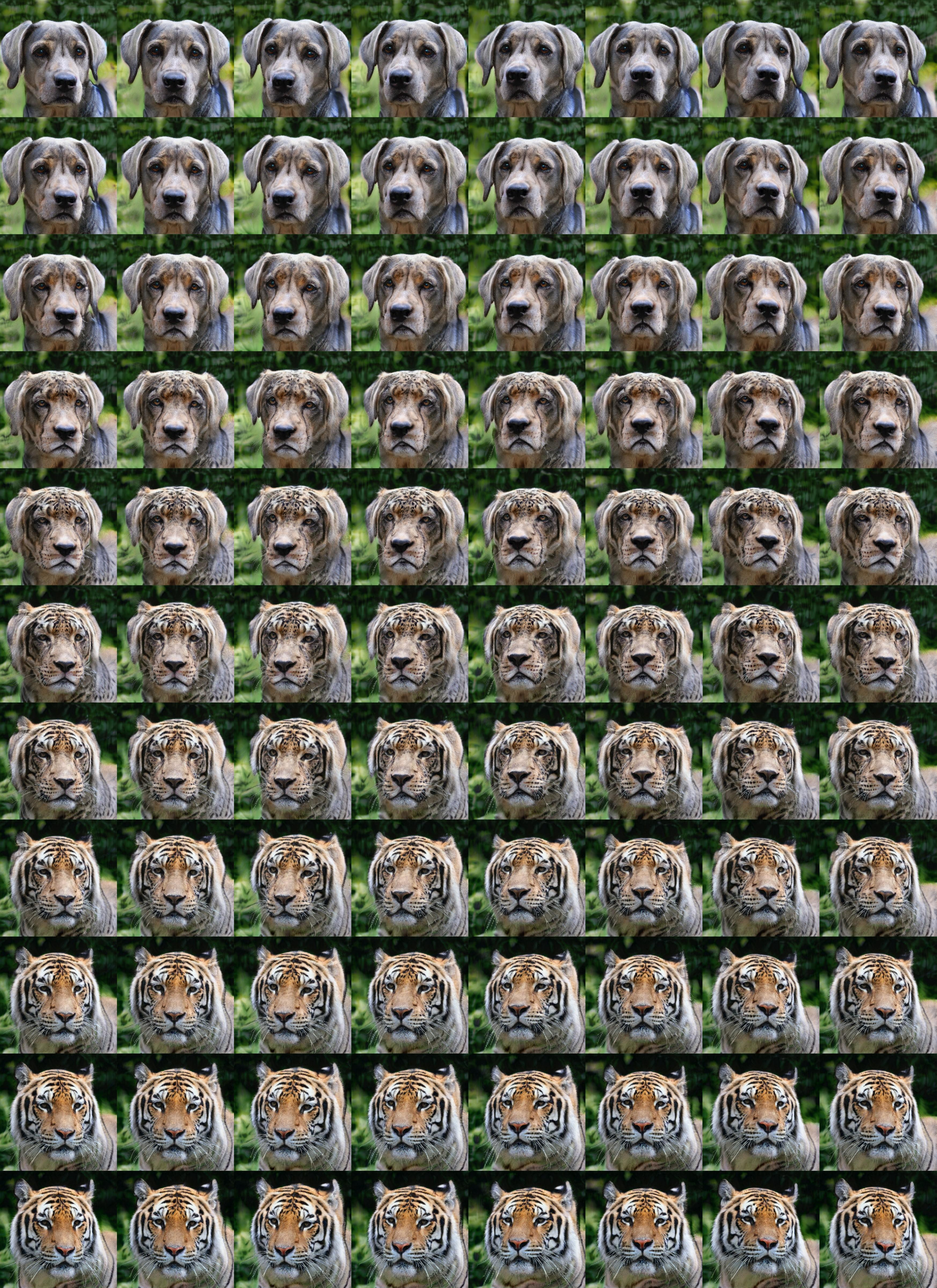}
    \captionof{figure}{Interpolation between dog and wild on AFHQ $256^2$.}  \vspace{-0.2 cm}
    \label{fig:inter1}
\end{center}
}]

\twocolumn[{
\renewcommand\twocolumn[1][]{#1}
\maketitle
\vspace{-0.7 cm}
\begin{center}
    \centering 
    \vspace{-0.0 cm}
        \includegraphics[width=0.81\textwidth]{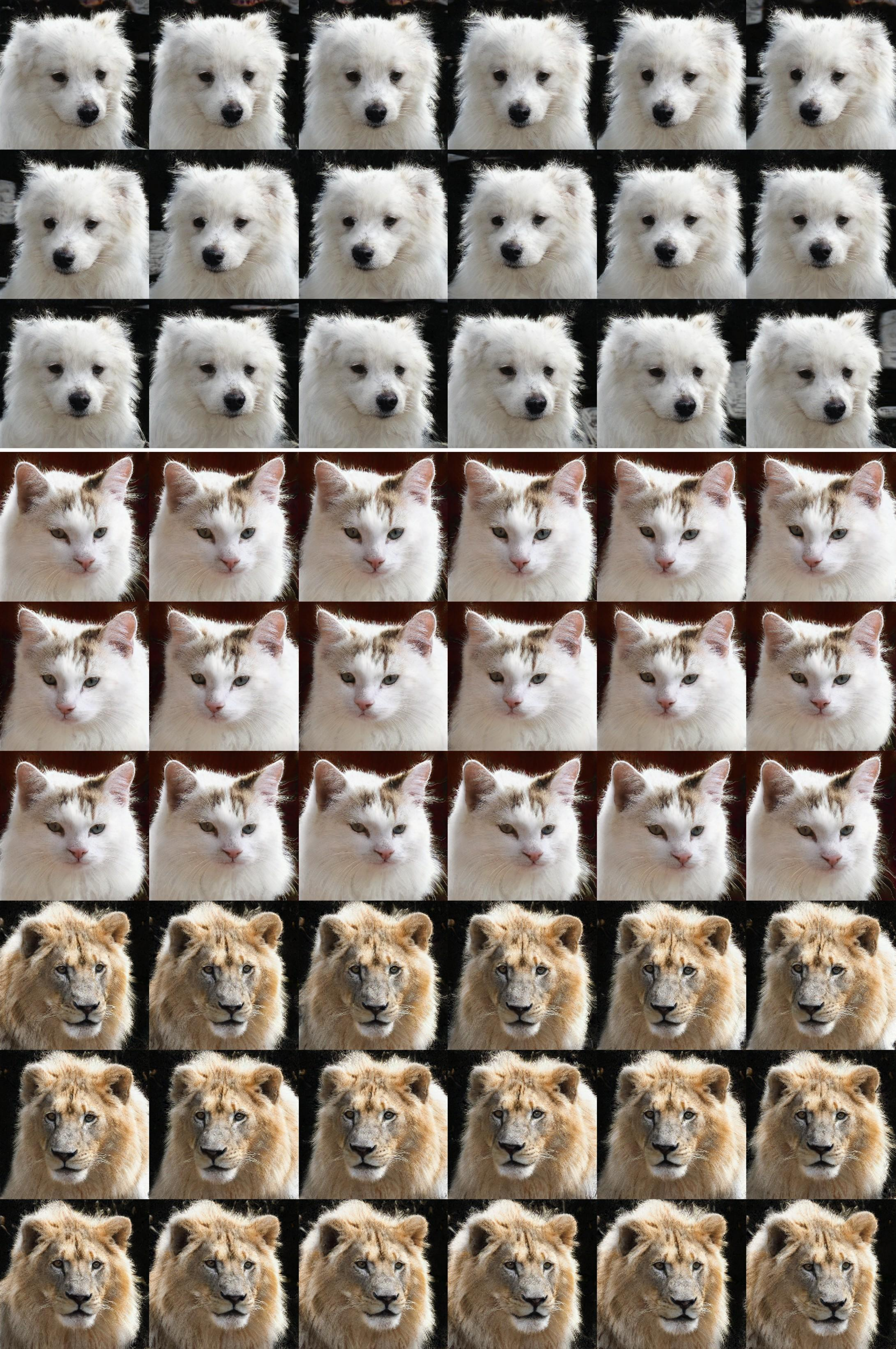}
    \captionof{figure}{Example of 3D-aware I2I translation of dog into cat and wild on AFHQ $256^2$.}  \vspace{-0.2 cm}
    \label{fig:add_afhq2}
\end{center}
}]

\twocolumn[{
\renewcommand\twocolumn[1][]{#1}
\maketitle
\vspace{-0.7 cm}
\begin{center}
    \centering 
    \vspace{-0.0 cm}
        \includegraphics[width=0.81\textwidth]{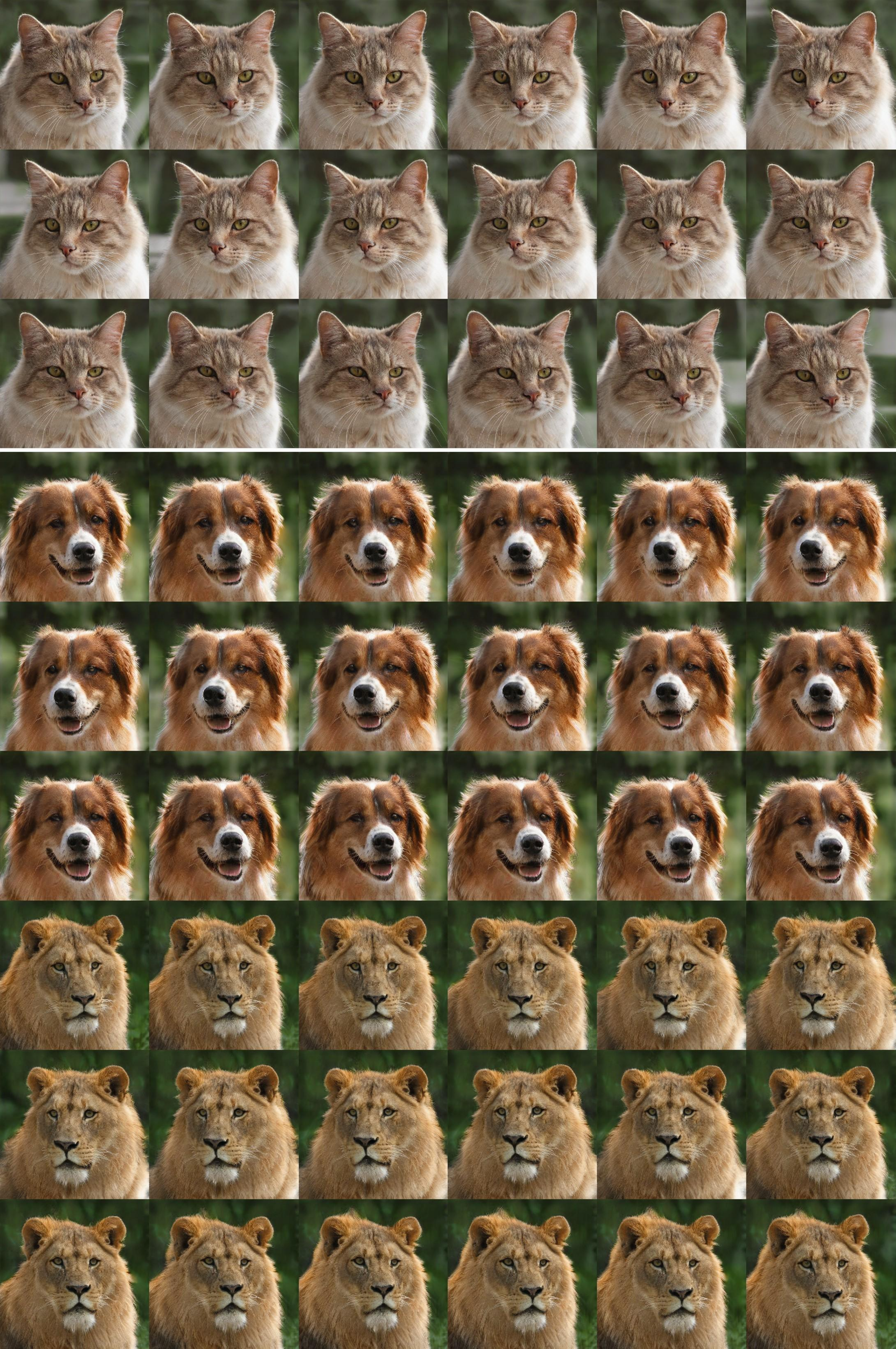}
    \captionof{figure}{Example of 3D-aware I2I translation of cat into dog and wild on AFHQ $256^2$.}  \vspace{-0.2 cm}
    \label{fig:add_afhq3}
\end{center}
}]

\twocolumn[{
\renewcommand\twocolumn[1][]{#1}
\maketitle
\vspace{-0.7 cm}
\begin{center}
    \centering 
    \vspace{-0.0 cm}
        \includegraphics[width=0.81\textwidth]{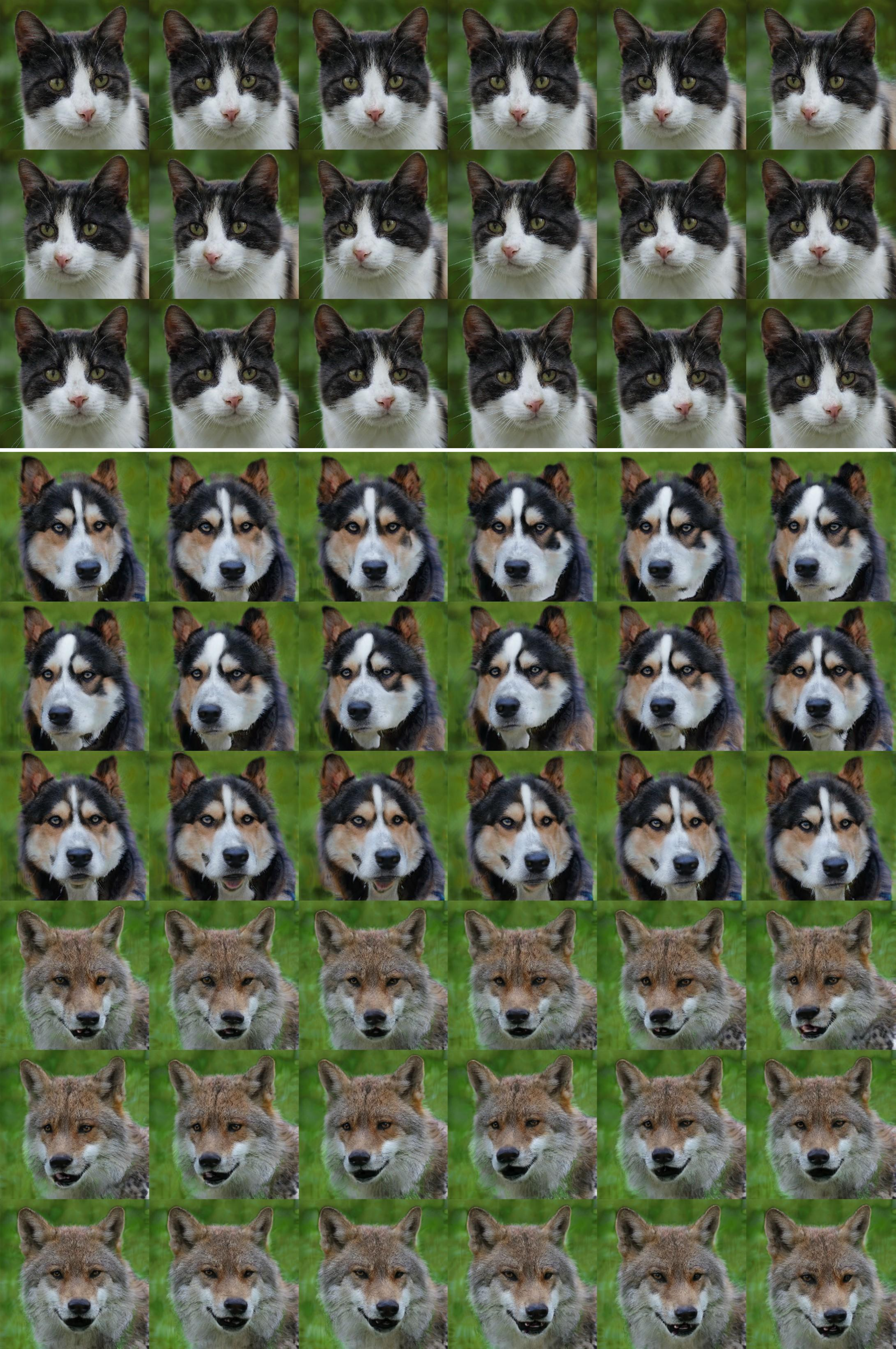}
    \captionof{figure}{Example of 3D-aware I2I translation of cat into dog and wild on AFHQ $256^2$.}  \vspace{-0.2 cm}
    \label{fig:add_afhq4}
\end{center}
}]

\twocolumn[{
\renewcommand\twocolumn[1][]{#1}
\maketitle
\vspace{-0.7 cm}
\begin{center}
    \centering 
    \vspace{-0.0 cm}
        \includegraphics[width=0.81\textwidth]{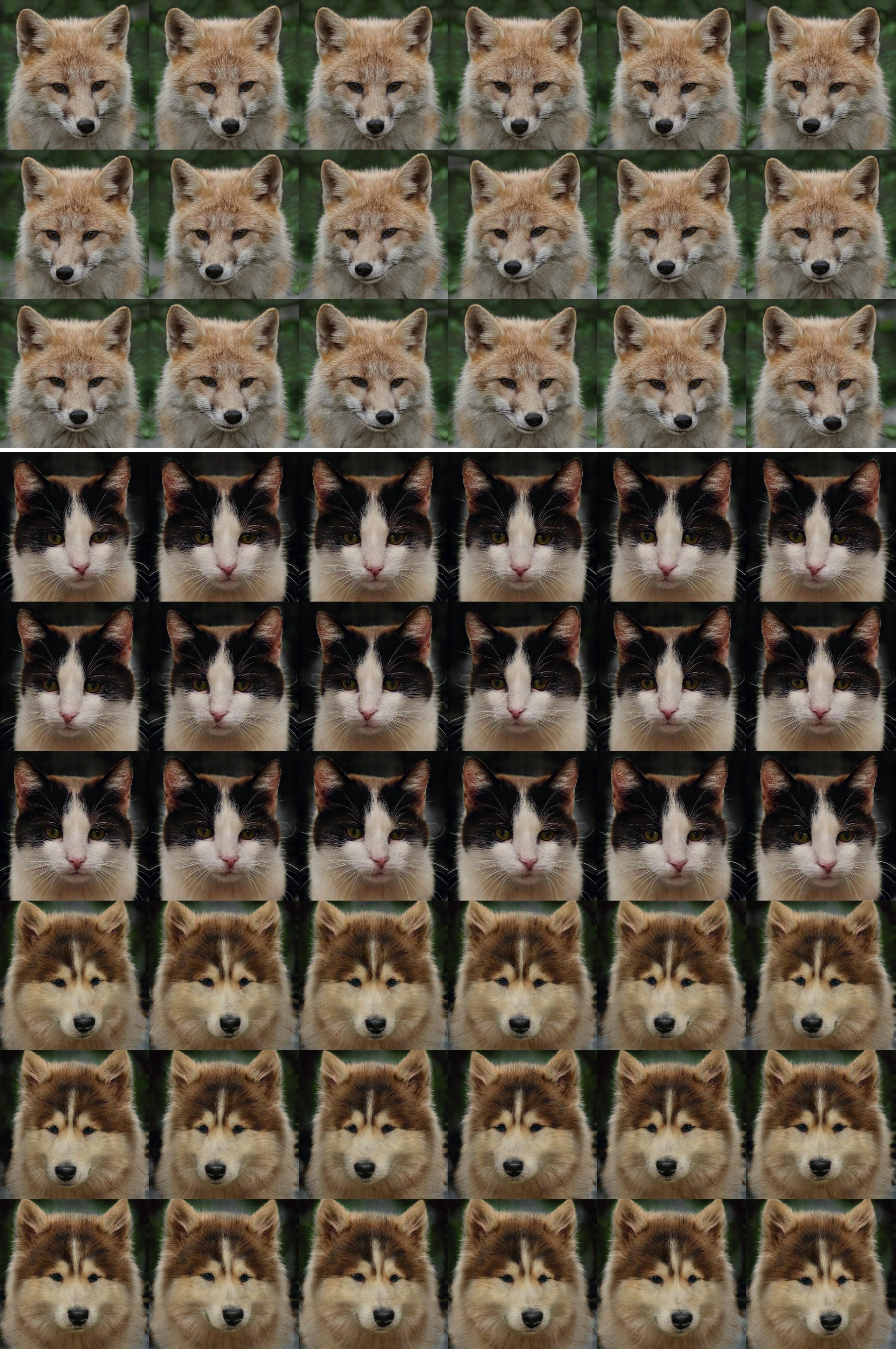}
    \captionof{figure}{Example of 3D-aware I2I translation of wild into cat and dog on AFHQ $256^2$.}  \vspace{-0.2 cm}
    \label{fig:add_afhq5}
\end{center}
}]

\twocolumn[{
\renewcommand\twocolumn[1][]{#1}
\maketitle
\vspace{-0.7 cm}
\begin{center}
    \centering 
    \vspace{-0.0 cm}
        \includegraphics[width=0.81\textwidth]{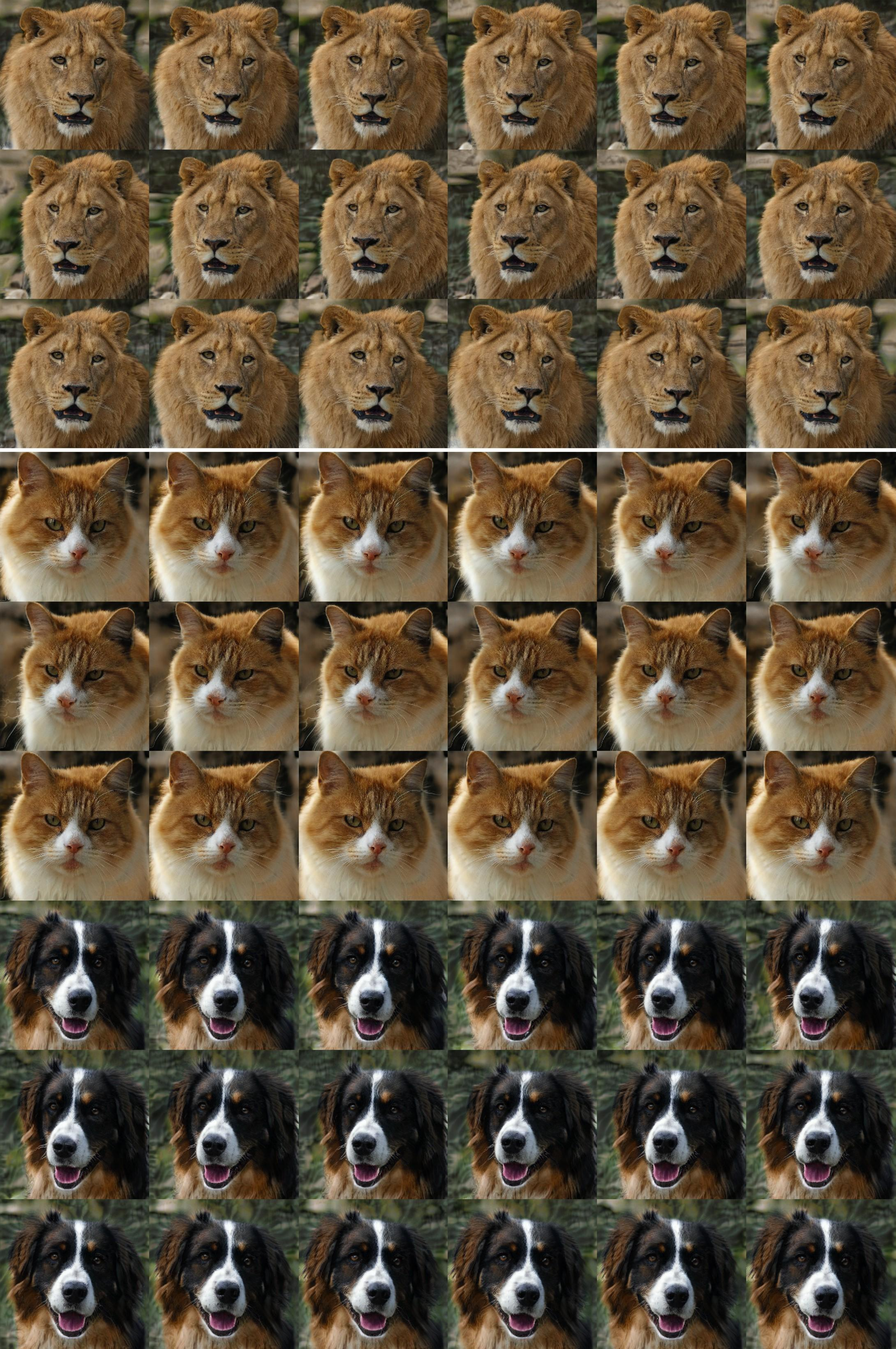}
    \captionof{figure}{Example of 3D-aware I2I translation of wild into cat and dog on AFHQ $256^2$.}  \vspace{-0.2 cm}
    \label{fig:add_afhq6}
\end{center}
}]

\twocolumn[{
\renewcommand\twocolumn[1][]{#1}
\maketitle
\vspace{-0.7 cm}
\begin{center}
    \centering 
    \vspace{-0.0 cm}
        \includegraphics[width=0.85\textwidth]{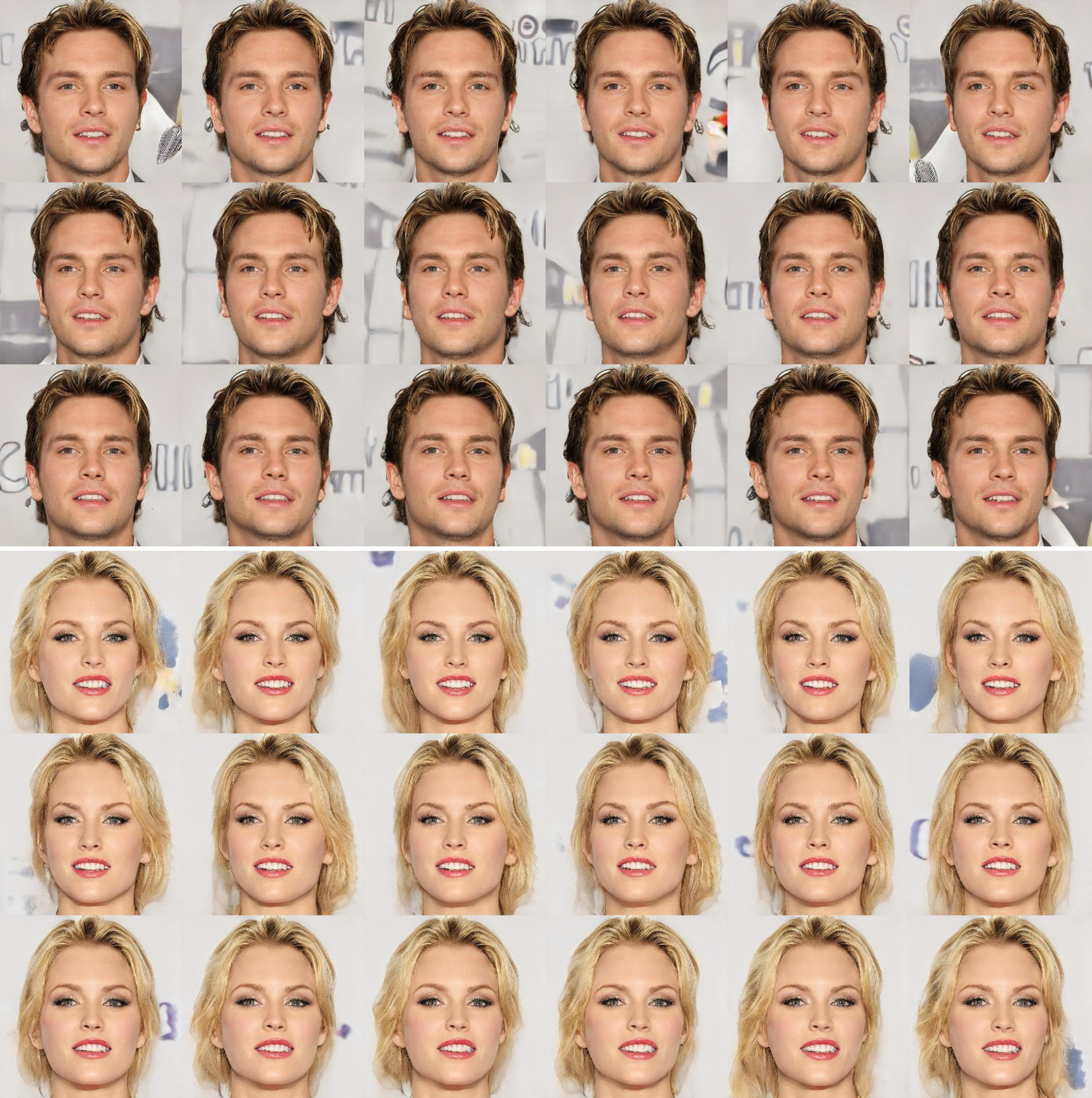}
    \captionof{figure}{Example of 3D-aware I2I translation of male into female on Celeba-HQ $256^2$.}  \vspace{-0.2 cm}
    \label{fig:add_celeba_hq1}
\end{center}
}]

\twocolumn[{
\renewcommand\twocolumn[1][]{#1}
\maketitle
\vspace{-0.7 cm}
\begin{center}
    \centering 
    \vspace{-0.0 cm}
        \includegraphics[width=0.85\textwidth]{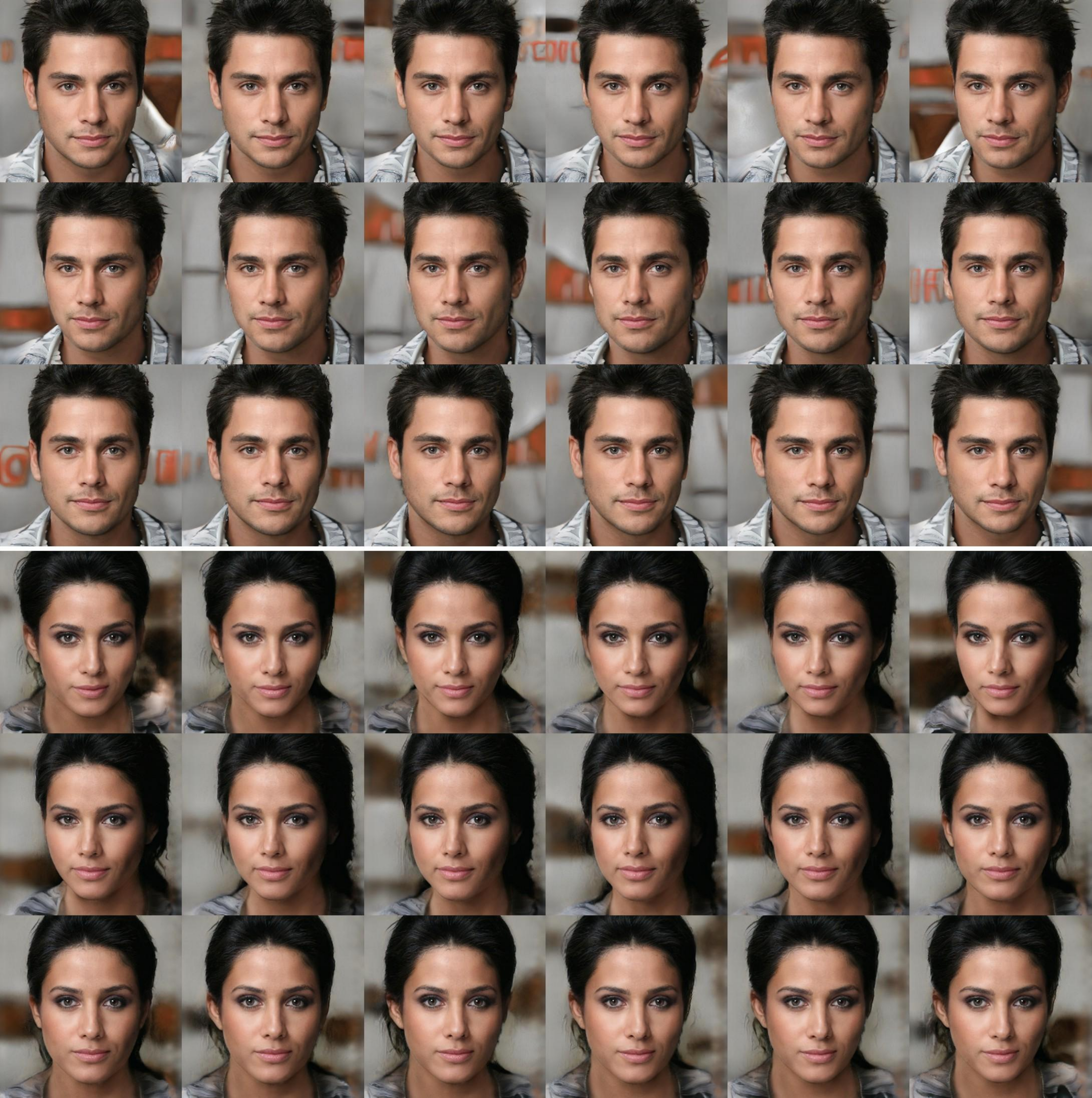}
    \captionof{figure}{Example of 3D-aware I2I translation of male into female on Celeba-HQ $256^2$.}  \vspace{-0.2 cm}
    \label{fig:add_celeba_hq2}
\end{center}
}]

\twocolumn[{
\renewcommand\twocolumn[1][]{#1}
\maketitle
\vspace{-0.7 cm}
\begin{center}
    \centering
    \vspace{-0.0 cm}
        \includegraphics[width=0.85\textwidth]{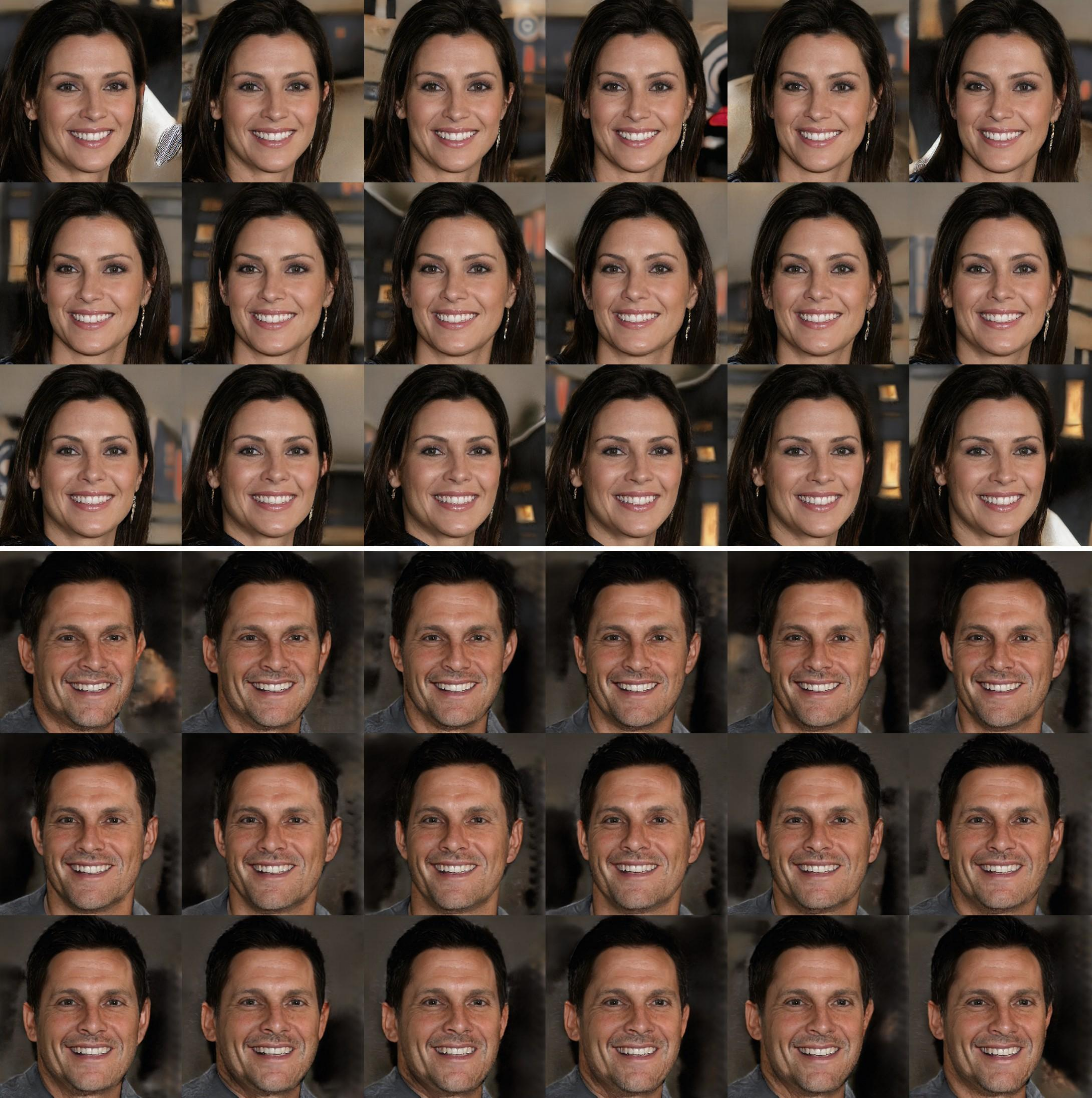}
    \captionof{figure}{Example of 3D-aware I2I translation of female into male on Celeba-HQ $256^2$.}  \vspace{-0.2 cm}
    \label{fig:add_celeba_hq3}
\end{center}
}]

\twocolumn[{
\renewcommand\twocolumn[1][]{#1}
\maketitle
\vspace{-0.7 cm}
\begin{center}
    \centering
    \vspace{-0.0 cm}
        \includegraphics[width=0.85\textwidth]{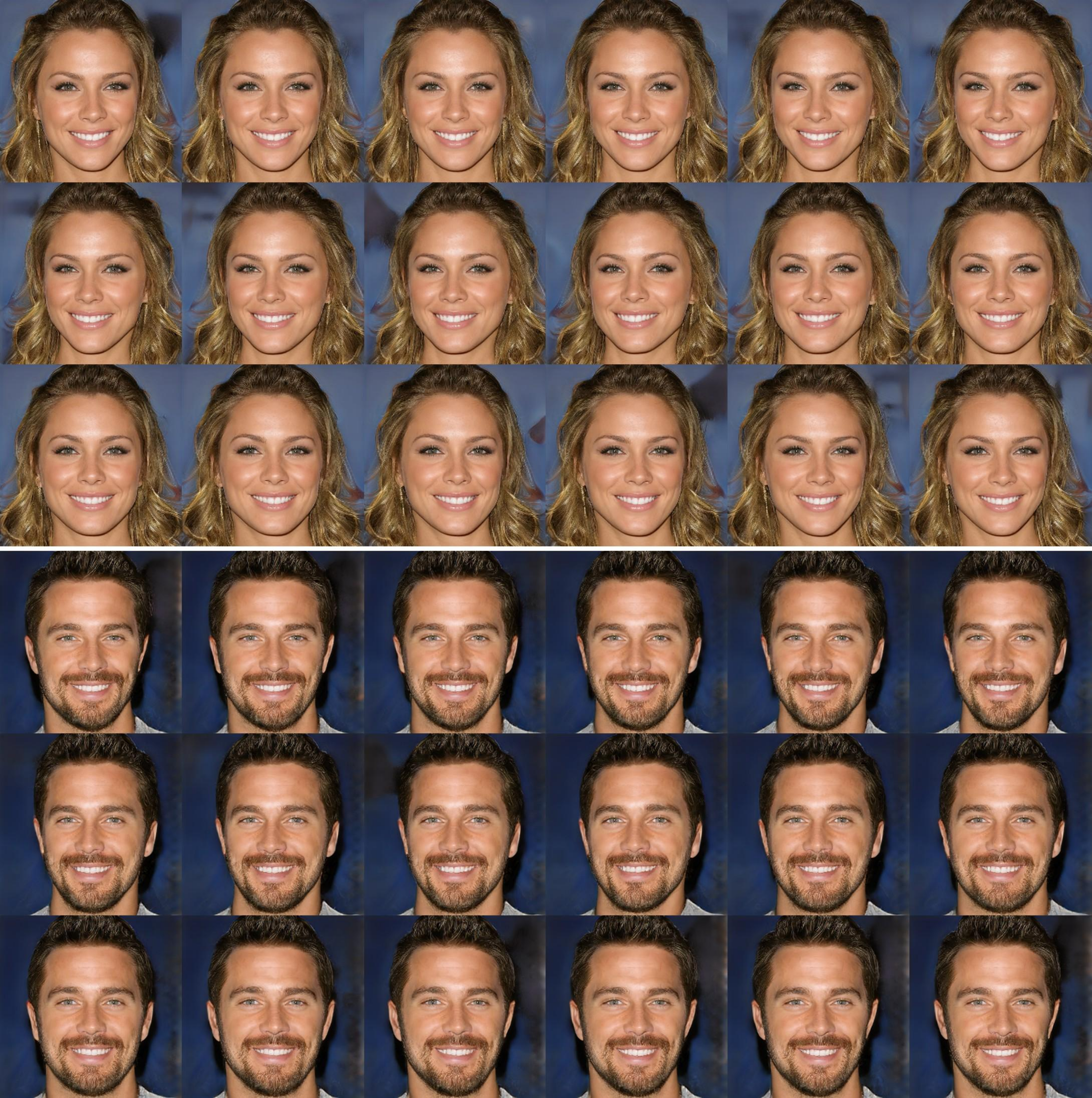}
    \captionof{figure}{Example of 3D-aware I2I translation of female into male on Celeba-HQ $256^2$.}  \vspace{-0.2 cm}
    \label{fig:add_celeba_hq4}
\end{center}
}]

\twocolumn[{
\renewcommand\twocolumn[1][]{#1}
\maketitle
\vspace{-0.7 cm}
\begin{center}
    \centering
    \vspace{-0.0 cm}
        \includegraphics[width=0.95\textwidth]{suppimgs/Figure18.pdf}
    \captionof{figure}{Example of 3D-aware I2I translation of female into male on Celeba-HQ $1024^2$.}  \vspace{-0.2 cm}
    \label{fig:celeba1024_3}
\end{center}
}]

\twocolumn[{
\renewcommand\twocolumn[1][]{#1}
\maketitle
\vspace{-0.7 cm}
\begin{center}
    \centering
    \vspace{-0.0 cm}
        \includegraphics[width=0.9\textwidth]{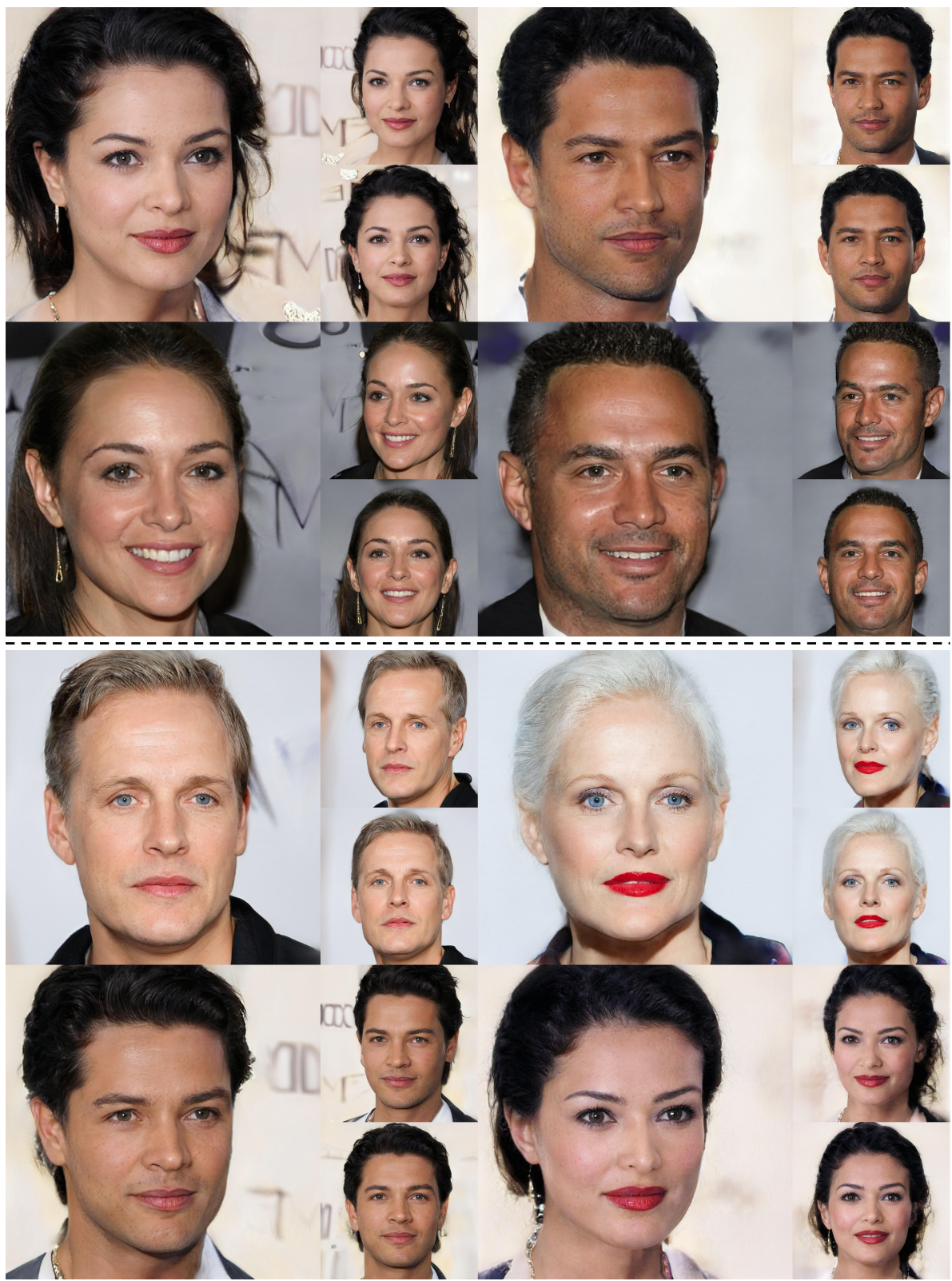}
    \captionof{figure}{Example of 3D-aware I2I translation of female into male (top) and male into female (bottom) on Celeba-HQ $1024^2$.}  \vspace{-0.2 cm}
    \label{fig:celeba1024_4}
\end{center}
}]

\end{document}